\documentclass{article}

\PassOptionsToPackage{numbers,  compress}{natbib} 
     \usepackage[final]{neurips_2023}

\usepackage[utf8]{inputenc} 
\usepackage[T1]{fontenc}    

\usepackage{xcolor}         
\definecolor{refcolor}{rgb}{0.0, 0.0, 0.85}

\RequirePackage[colorlinks,linkcolor=refcolor,citecolor=refcolor,urlcolor=refcolor,linktoc=page]{hyperref}       

\usepackage{url}            
\usepackage{booktabs}       
\usepackage{amsfonts}       
\usepackage{nicefrac}       
\usepackage{microtype}      
\usepackage{tikz}
\usepackage{amsmath}
\usepackage{comment}
\usepackage{cancel}
\usepackage{bbm}

\usetikzlibrary{cd,arrows,calc,shapes,positioning}
\tikzstyle{obs} = [circle,fill=white,draw=black,inner sep=0pt,minimum size=18pt,font=\fontsize{10}{10}\selectfont,node distance=1,thick]

\tikzstyle{obsdummy} = [circle,fill opacity=0,draw=none,inner sep=0pt,minimum size=18pt,font=\fontsize{10}{10}\selectfont,node distance=1,thick]

\tikzstyle{hobs} = [circle,fill=white,draw=black,inner sep=0pt,minimum size=18pt,font=\fontsize{10}{10}\selectfont,node distance=1,ultra thick]
\tikzstyle{intv} = [rectangle,fill=white,draw=black,inner sep=0pt,minimum size=18pt,font=\fontsize{10}{10}\selectfont,node distance=1,thick]
\tikzstyle{hintv} = [rectangle,fill=white,draw=white,inner sep=0pt,minimum size=18pt,font=\fontsize{10}{10}\selectfont,node distance=1,ultra thick]
\tikzstyle{fact} = [rectangle,fill=black,draw=black,inner sep=0pt,minimum size=10pt,font=\fontsize{10}{10}\selectfont,node distance=1,thick]
\tikzstyle{latent} = [circle,fill=white,draw=black,inner sep=0pt,minimum size=18pt,font=\fontsize{10}{10}\selectfont,node distance=1,thick,dotted]

\newcommand{\edge}[3][]{ %
  \foreach \x in {#2} { %
    \foreach \y in {#3} { %
      \path (\x) edge [->, >={Latex[round]}, #1,thick] (\y) ;%
    } ;
  } ;
}
\newcommand{\uedge}[3][]{ %
  \foreach \x in {#2} { %
    \foreach \y in {#3} { %
      \path (\x) edge [-, >={Latex[round]}, #1,thick] (\y) ;%
    } ;
  } ;
}
\newcommand{\bedge}[3][]{ %
  \foreach \x in {#2} { %
    \foreach \y in {#3} { %
      \path (\x) edge [<->, >={Latex[round]}, #1,thick] (\y) ;%
    } ;
  } ;
}

\definecolor{lightgrey}{rgb}{0.925, 0.925, 0.925}

\newcommand{\train}{\mathrm{train}}
\newcommand{\test}{\mathrm{test}}

\DeclareMathOperator{\indep}{\perp\!\!\!\perp}

\newtheorem{theorem}{Theorem}[section]

\newtheorem{lemma}[theorem]{Lemma}

\title{Intervention Generalization:\\A View from Factor Graph Models}

\author{%
  Gecia Bravo-Hermsdorff\thanks{Department of Statistical Science, University College London.}\\
  \And
  David S. Watson\thanks{Department of Informatics, King's College London.}\\
  \And
  Jialin Yu\footnotemark[1]\\\\
  \And
  Jakob Zeitler\thanks{Department of Computer Science, University College London.}\\
  \And
  Ricardo Silva\footnotemark[1]\\
}

\begin{document}

\maketitle

\begin{abstract}

One of the goals of causal inference is to generalize from past experiments and observational data to novel conditions. 
While it is in principle possible to eventually learn a mapping from a novel experimental condition to an outcome of interest, provided a sufficient variety of experiments is available in the training data, coping with a large combinatorial space of possible interventions is hard. 
Under a typical sparse experimental design, this mapping is \mbox{ill-posed} without relying on heavy regularization or prior distributions. 
Such assumptions may or may not be reliable, and can be hard to defend or test.
In this paper, we take a close look at how to warrant a leap from past experiments to novel conditions based on minimal assumptions about the factorization of the distribution of the manipulated system, communicated in the \mbox{well-understood} language of factor graph models. 
A postulated \emph{interventional factor model} (IFM) may not always be informative, but it conveniently abstracts away a need for 
explicitly modeling unmeasured confounding and feedback mechanisms, leading to directly testable claims. 
Given an IFM and datasets from a collection of experimental regimes, we derive conditions for identifiability of the expected outcomes of new regimes never observed in these training data. 
We implement our framework using several efficient algorithms, and apply them on a range of \mbox{semi-synthetic} experiments. 
\end{abstract}

\section{Introduction}
Causal inference is a fundamental problem in many sciences, such as clinical medicine \cite{bica2021, squires2022causal,shi2022assumptions} and molecular biology \cite{sachs2005causal, Karlebach2008, gentzel2021and}. 
For example, causal inference can be used to identify the effects of chemical compounds on cell types \cite{squires2022causal} or determine the underlying mechanisms of disease \cite{leist2022mapping}. 

One particular challenge in causal inference is \textit{generalization} --- the ability to extrapolate knowledge gained from past experiments and observational data to previously unseen scenarios. 
Consider a laboratory that has performed several gene knockouts and recorded subsequent outcomes. 
Do they have sufficient information to predict how the system will behave under some new combination(s) of knockouts?
Conducting all possible experiments in this setting would be prohibitively expensive and time consuming.
A supervised learning method could, in principle, map a vector representation of the design to outcome variables of interest. 
However, past experimental conditions may be too sparsely distributed in the set of all possible assignments, and such a direct supervised mapping would require leaps of faith about how assignment decisions interact with the outcome, even if under the guise of formal assumptions such as linearity. 


In this paper, we propose a novel approach to the task of \textit{intervention generalization}, i.e., predicting the effect of unseen treatment regimes.  
We rely on little more than a postulated factorization of the distribution describing how intervention variables $\sigma$ interact with a random process $X$, and a possible target outcome $Y$. 

Related setups appear in the causal modeling literature on soft interventions \citep{correa_completeness_sigma}, including its use in methods such as causal bandits \citep{lattimore:2016} and causal Bayesian optimization \citep{aglietti:20} (see Section \ref{sec:related}). 
However, these methods rely on directed acyclic graphs (DAGs), which may be hard to justify in many applications. 
Moreover, in the realistic situation where the target of a \mbox{real-world} intervention is a \emph{set} of variables \citep{eaton:07} and not only the ``child'' variable within a DAG factor, the \mbox{parent-child} distinction is blurred and previous identification results do not apply. 
%
Without claiming that our proposal is appropriate for every application, we suggest the following take on intervention generalization: \emph{model a causal structure as a set of soft constraints, together with their putative (arbitrary but local) modifications by external actions}.
Our \textit{interventional factor model} (IFM) leads to provable intervention generalization via a factor graph decomposition which, when informative, can be tested without further assumptions beyond basic relations of conditional independence. 
IFMs are fully agnostic with regards to cycles or hidden variables, 
in the spirit of \citet{dawid:21}'s \mbox{decision-theoretic} approach to causal inference, 
where the key ingredient boils down to statements of conditional independence among random and intervention variables.

Our primary contributions are as follows. 
(1) We introduce the \textit{interventional factor model} (IFM), which aims to solve intervention generalization problems using only claims about which interventions interact with which observable random variables. 
(2) We establish necessary and sufficient conditions for the identifiability of treatment effects within the IFM framework. 
(3) We adapt existing results from conformal inference to our setting, providing \mbox{distribution-free} predictive intervals for identifiable causal effects with guaranteed finite sample coverage. 
(4) We implement our model using efficient algorithms, and apply them to a range of \mbox{semi-synthetic} experiments.

\section{Problem Statement}
\label{sec:prob}
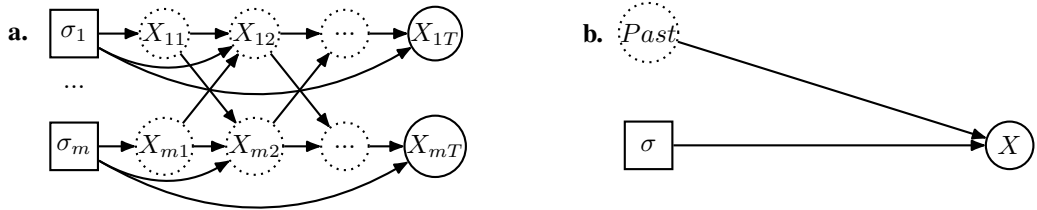
\begin{figure}[!t]
  \begin{tabular}{cc}
  \begin{tikzpicture}
    \node[right] (L) at (-1, 3) {\textbf{a.}};
    \node[latent] (X11) at (1.2, 3) {$X_{11}$};
    \node[latent] (X21) at (1.2, 1.5) {$X_{m1}$};
    \node[latent] (X12) at (2.4, 3) {$X_{12}$};
    \node[latent] (X22) at (2.4, 1.5) {$X_{m2}$};
    \node[latent] (G1) at (3.6, 3) {$...$};
    \node[latent] (G2) at (3.6, 1.5) {$...$};
    \node[obs] (X1t) at (4.8, 3) {$X_{1T}$};
    \node[obs] (X2t) at (4.8, 1.5) {$X_{mT}$};
    \node[intv] (S1) at (0, 3) {$\sigma_1$};
    \node[obs, draw=none] (Sd) at (0, 2.3){$...$};
    \node[intv] (S2) at (0, 1.5) {$\sigma_m$};
    \edge[black]{S1}{X11};
    \edge[black]{S2}{X21};
    \edge[black]{X11}{X12};
    \edge[black]{X11}{X22};
    \edge[black]{X21}{X12};
    \edge[black]{X21}{X22};
    \edge[black]{X12}{G1};
    \edge[black]{X12}{G2};
    \edge[black]{X22}{G1};
    \edge[black]{X22}{G2};
    \edge[black]{G1}{X1t};
    \edge[black]{G2}{X2t};
    \edge[->, bend right=30]{S1}{X12};
    \edge[->, bend right=30]{S2}{X22};
    \edge[->, bend right=30]{S1}{X1t};
    \edge[->, bend right=30]{S2}{X2t};
  \end{tikzpicture}%
  \hspace*{1.0cm}
  & 
  \begin{tikzpicture}
    \node[right] (L) at (-1, 3) {\textbf{b.}};
    \node[intv] (S) at (0, 1.5) {$\sigma$};
    \node[obsdummy] (Sd) at (0, 0.8) {dum};
    \node[latent] (P) at (0, 3) {$Past$};
    \node[obs] (X) at (4.8, 1.5) {$X$};
    \edge[black]{P}{X};
    \edge[black]{S}{X};
    \vspace{5in}
 \end{tikzpicture}%
 \end{tabular}
 \vspace{-4mm}
  \caption{
  \textbf{(a)} A dynamic process where only the final step is (simultaneously) observed. 
   Intervention variables are represented as white squares, random variables as circles, and dashed lines indicate hidden variables. 
  \textbf{(b)} A bird's eye unstructured view of the sampling process, where an intervention vector dictates $\sigma_{ }^{ }$ the distribution obtained at an \mbox{agreed-upon} time frame $T$.
  }
 \label{fig:process}
 \vspace{-2mm}
\end{figure}

\paragraph{Data assumptions.} 
Fig.~\ref{fig:process}\textbf{(a)} illustrates a generative process common to many applications and particularly suitable to our framework: 
a perturbation, here represented by a set of interventional variables $\sigma$, is applied and a dynamic feedback loop takes place. 
Often in these applications (e.g, experiments in cell biology \citep{sachs2005causal} and social science \citep{ogburn:20}), the sampling of this process is highly restrictive, and the \mbox{time-resolution} may boil down to a single \emph{snapshot}, as illustrated by Fig.~\ref{fig:process}\textbf{(b)}. 
We assume that data is given as samples of a random vector $X$ collected \mbox{cross-sectionally} at a \mbox{well-defined} \mbox{time-point} (not necessarily at equilibrium) under a \mbox{well-defined} intervention vector $\sigma$.
The importance of establishing a clear sampling time in the context of graphical causal models is discussed by \citep{dash:05}.
%
The process illustrated in Fig.~\ref{fig:process}\textbf{(a)} may suggest no particular Markovian structure at the time the snapshot is taken. 
However, in practice, it is possible to model the \mbox{black-box} sampling process represented in Fig.~\ref{fig:process}\textbf{(b)} in terms of an \emph{energy function} representing soft constraints \citep{lecun:06}. 
These could be the result of particular equilibrium processes \citep{lauritzen:02}, including deterministic differential equations \citep{iwasaki:94}, or \mbox{empirically-verifiable} approximations \citep{ogburn:20}. 

\emph{The role of a causal model is to describe how intervention $\sigma$ locally changes the energy function.} 
In the context of causal DAG models, sometimes this takes the guise of ``soft interventions'' and other variations that can be called ``interventions on structure'' \citep{korb:04, malinksy:18} or edge interventions \citep{shptiser:16a}. 
Changes of structural coefficients in possibly cyclic models have also been considered \citep{hoover:01}. 
The model family we propose takes this to the most abstract level, modeling energy functions via recombinations of local interventional effects without acyclicity constraints, as such constraints should not be taken for granted \citep{dawid:10} and are at best a crude approximation for some problems at hand (e.g., \citep{sachs2005causal}).

\begin{figure}[!t]
   \raisebox{0.89cm}{
  \hspace*{0.1cm}
  \centering
  \begin{tikzpicture}
    \node[right] (L) at (-1, 2) {\textbf{a.}};
    \node[obs] (X1) at (0, 0) {$X_1$};
    \node[obs] (X2) at (1.2, 0) {$X_2$};
    \node[obs] (X3) at (2.4, 0) {$X_3$};
    \node[intv] (S1) at (0, 1) {$\sigma_1$};
    \node[intv] (S2) at (1.2, 1) {$\sigma_2$};
    \node[intv] (S3) at (2.4, 1) {$\sigma_3$};
    \edge[black]{X1}{X2};
    \edge[black]{X2}{X3};
    \edge[->, bend right=30]{X1}{X3};
    \edge[black]{S1}{X1};
    \edge[black]{S2}{X2};
    \edge[black]{S3}{X3};
  \end{tikzpicture}%
  \hspace*{1.0cm}
  \begin{tikzpicture}
    \node[right] (L) at (-1, 2) {\textbf{b.}};
    \node[obs] (X1) at (0, 0) {$X_1$};
    \node[obs] (X2) at (1.2, 0) {$X_2$};
    \node[obs] (X3) at (2.4, 0) {$X_3$};
    \node[fact] (F1) at (0, 1) {\textcolor{white}{$f_1$}};
    \node[fact] (F2) at (1.2, 1) {\textcolor{white}{$f_2$}};;
    \node[fact] (F3) at (2.4, 1) {\textcolor{white}{$f_3$}};;    
    \node[intv] (S1) at (0, 2) {$\sigma_1$};
    \node[intv] (S2) at (1.2, 2) {$\sigma_2$};
    \node[intv] (S3) at (2.4, 2) {$\sigma_3$};
    \edge[black]{S1}{F1};
    \uedge[black]{X1}{F1};
    \edge[black]{S2}{F2};
    \uedge[black]{X1}{F2};
    \uedge[black]{X2}{F2};
    \edge[black]{S3}{F3};
    \uedge[black]{X1}{F3};
    \uedge[black]{X2}{F3};
    \uedge[black]{X3}{F3};
  \end{tikzpicture}%
  \hspace*{1.0cm}
  \begin{tikzpicture}
    \node[right] (L) at (-1, 2) {\textbf{c.}};
    \node[obs] (X1) at (0, 0) {$X_1$};
    \node[obs] (X2) at (1.2, 0) {$X_2$};
    \node[obs] (X3) at (2.4, 0) {$X_3$};
    \node[intv] (S1) at (0, 1) {$\sigma_1$};
    \node[intv] (S2) at (1.2, 1) {$\sigma_2$};
    \node[intv] (S3) at (2.4, 1) {$\sigma_3$};
    \uedge[black]{X1}{X2};
    \uedge[black]{X2}{X3};
    \edge[-, bend right=30]{X1}{X3};
    \edge[black]{S1}{X1};
    \edge[black]{S2}{X2};
    \edge[black]{S2}{X1};
    \edge[black]{S3}{X3};
    \edge[black]{S3}{X2};
    \edge[black]{S3}{X1};
  \end{tikzpicture}%
 }\\[-5mm]
 \vspace{-5mm}
 \caption{
 Examples of distinct causal graphical models, expressing different factorization assumptions. 
 Random variables are represented as circles, intervention variables as white squares, and factors as black squares. 
  \textbf{(a)} A directed acyclic graph (DAG) with explicit intervention variables. 
  \textbf{(b)} The corresponding interventional factor model (IFM) for \textbf{(a)}. 
  \textbf{(c)} A Markov random field (MRF) with interventional variables, thus, forming a chain graph. 
 }
 \label{fig:graph_differences}
 \vspace{-2mm}
\end{figure}
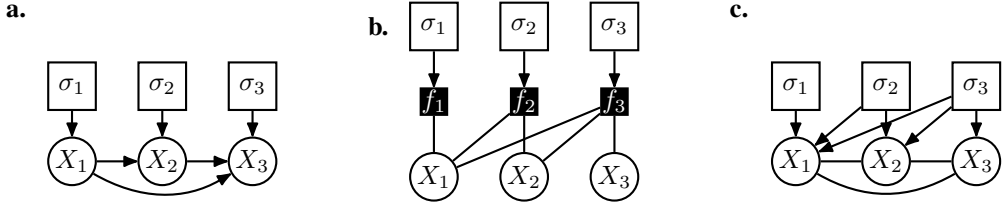

\paragraph{Background and notation.} 
The notion of an \emph{intervention variable} in causal inference encodes an action that modifies the distribution of a system of random variables. 
This notion is sometimes brought up explicitly in graphical formulations of causal models \citep{pearl:09, sgs:00, dawid:21}. 
To formalize it, let $\sigma$ denote a vector of intervention variables (also known as \textit{regime indicators}), with each $\sigma_i$ taking values in a finite set \mbox{$\{0, 1, 2, \dots, \aleph_i - 1\}$}. 
A fully specified vector of intervention variables characterizes a \textit{regime} or an \textit{environment} (we use these terms interchangeably). 
Graphically, we represent intervention variables using white squares and random variables using circles (Fig.~\ref{fig:graph_differences}). 
We use subscripts to index individual (random or interventional) variables and superscripts to index regimes. 
For instance, $X_i^j$ denotes the random variable $X_i$ under regime $\sigma^j$, while $\sigma_i^j$ denotes the $i$th intervention variable of the $j$th environment. (Note that the order of the variables is arbitrary.) 
We occasionally require parenthetical superscripts to index samples, so \smash{$x_i^{j(k)}$} denotes the $k$th sample of $X_i$ under regime $\sigma^j$.

For causal DAGs,
Pearl's $do$ operator \citep{pearl:09} denotes whether a particular random variable $X_i$ is manipulated to have a given value (regardless of the values of its parents). 
As an example, consider a binary variable \mbox{$X_i \in \{\textsc{True}, \textsc{False}\}$}.  
We can use a (categorical) intervention variable $\sigma_i$ to index the two ``do-interventional'' regimes: 
\mbox{$\sigma_i = 1$} denotes ``\mbox{$do(X_i = \textsc{True})$}'', and \mbox{$\sigma_i = 2$} denotes ``\mbox{$do(X_i = \textsc{False})$}.'' 
We reserve $\sigma_i = 0$ to denote the choice of ``no manipulation'', also known as the ``observational'' regime.  
However, one need not think of this \mbox{$\sigma_i = 0$} setting as fundamentally different than the others; indeed, it is convenient to treat all regimes as choices of data generating process.\footnote{In general, there can be infinitely many choices for interventional settings. 
However, for the identifiability results in the next section, we cover the finite case only, as it removes the need for smoothness assumptions on the effect of intervention levels.} 
Moreover, there is no need for intervention variables to correspond to deterministic settings of random variables; they may in principle describe any \mbox{well-defined} change in distribution, such as stochastic or conditional interventions \citep{dawid:21,correa_sigma_calc}.  
As a stochastic example, \mbox{$\sigma_i = 3$} could correspond to randomly choosing \mbox{$do(X_i=\textsc{True})$} 30\% of the time and \mbox{$do(X_i=\textsc{False})$} 70\% of the time.  
As a conditional example, \mbox{$\sigma_i = 4$} could mean ``if parents of $X_i$ satisfy a given condition, \mbox{$do(X_i=\textsc{True})$}, otherwise do nothing''. 
When $X_i$ can take more values, the options for interventional regimes become even more varied.  
The main point to remember is that \mbox{$\sigma_i = 1, \sigma_i = 2, \ldots$} should be treated as distinct categorical options, and we reserve \mbox{$\sigma_i = 0$} to denote the ``observational'' case of ``no manipulation''.  


As discussed in the previous section, 
the probability density/mass function 
$p(x; \sigma)$ can be the result of a feedback process that does not naturally fit a DAG representation.  
Indeed, a growing literature in causal inference carefully considers how DAGs may give rise to equilibrium distributions (e.g., \citep{lauritzen:02,bongers:21,blom:23}), or marginals of continuous-time processes (e.g., \citep{mogensen:18}). 
However, they come with considerable added complexity of assumptions to ensure identifiability.
%
In this work, we abstract away all low-level details about how an equilibrium distribution comes to be, and instead require solely a model for how a distribution $p(x)$ factorizes as a function of $\sigma$. 
These assumptions are naturally formulated as a factor graph model \citep{kschis:01} augmented with intervention variables, which we call an \emph{interventional factor model (IFM)}.


\paragraph{Problem statement.} 
We are given a space $\Sigma$ of possible values for an intervention vector $\sigma$ of dimension $d$, making \mbox{$|\Sigma| \leq \prod_{i = 1}^d \aleph_i$}. 
For a range of training regimes \mbox{$\Sigma_{\train} \subset \Sigma$}, we are  given collection of datasets \smash{\mbox{$\mathcal D^{1}, \mathcal D^{2}, \dots, \mathcal D^{t}$}}, with dataset \smash{$\mathcal D^j$} collected under environment/regime \mbox{$\sigma^j_{ } \in \Sigma_{\train}$}. 
The goal is to learn \mbox{$p(x; \sigma^\star)$}, and \mbox{$\mu(\sigma^\star) := \mathbb E[Y; \sigma^\star]$}, for all test regimes \mbox{$\sigma^\star \in \Sigma_{\test} = \Sigma \backslash \Sigma_{\train}$}. 

In each training dataset, we measure a sample of \mbox{post-treatment} i.i.d.~draws of some \mbox{$m$-dimensional} random vector $X$ (possibly with \mbox{$m\neq d$}, as there is no reason to always assume a \mbox{one-to-one} mapping between intervention and random variables), and optionally an extra outcome variable $Y$.
The data generating process $p(x; \sigma^j)$ is unknown. 
We assume \mbox{$Y \indep \sigma~|~X$} for simplicity\footnote{Otherwise, we can just define \mbox{$Y := X_m$}, and use the identification results in Section~\ref{sec:ifm_model} to check whether $Y$ can be predicted from $\sigma$.}, which holds automatically in cases where $Y$ is a known deterministic summary of $X$. 
We are also given a factorization of $p(x; \sigma)$,
\vspace{-7pt}
\begin{equation}
p(x; \sigma) \propto \prod_{k = 1}^l f_k(x_{S_k}; \sigma_{F_k}),\ \forall \sigma \in \Sigma,
\label{eq:ifm}
\vspace{-2pt}
\end{equation}
\noindent where \mbox{$S_k \subseteq [m]$} and \mbox{$F_k \subseteq [d]$} define the causal structure. 
The (positive) functions \mbox{$f_k(\cdot; \cdot)$} are unknown. 
The model $p(y~|~x)$ can also be unknown, depending on the problem.

As illustrated in Figs.~\ref{fig:graph_differences}\textbf{(a-b)}, such an intervention factor model (IFM) can also encode (a relaxation of) the structural constraints implied by DAG assumptions.  
In particular, note the \mbox{one-to-one} correspondence between the DAG factorization in Fig.~\ref{fig:graph_differences}\textbf{(a)} and the factors in the IFM in Fig.~\ref{fig:graph_differences}\textbf{(b)}: 
\mbox{$f_1(x_1; \sigma_1) := p(x_1; \sigma_1)$}, \mbox{$f_2(x_1, x_2; \sigma_2) := p(x_2~|~x_1; \sigma_2)$}, and $f_3(x_1, x_2, x_3; \sigma_3) := p(x_3~|~x_1, x_2; \sigma_3)$. 
This explicitly represents that the conditional distribution of $x_1$ given all other 
variables fully factorizes in $\sigma$: i.e., $p(x_1~|~x_2, x_3; \sigma) \propto f_1(x_1; \sigma_1)f_2(x_1, x_2; \sigma_2)f_3(x_1, x_2, x_3; \sigma_3)$. 
Such factorization is not enforced by usual parameterizations of Markov random fields (i.e., graphical models with only undirected edges) or chain graphs (graphical models with acyclic directed edges and undirected edges) \citep{drton:09} such as the example shown in Fig.~\ref{fig:graph_differences}\textbf{(c)}.

\paragraph{Scope and limitations.} 
The factorization in Eq.~\eqref{eq:ifm} may come from different sources, e.g., from knowledge about physical connections (it is typically the case that one is able to postulate which variables are directly or only indirectly affected by an intervention),
or as the result of structure learning methods (e.g., \citep{abbeel:06}). 
For structure learning, \mbox{faithfulness-like} assumptions \citep{sgs:00} are required, as conditional independencies discovered under configurations $\Sigma_{\train}$ can only be extrapolated to $\Sigma_{\test}$ by assuming  that independencies observed over particular values of $\sigma$ can be generalized across all regimes. 
We do not commit to any particular structure learning technique, and refer to the literature on eliciting and learning graphical structure for a variety of methods \citep{koller:09}. 
In Appendix \ref{appendix:structure}, we provide a general guide on structure elicitation and learning, and a primer on causal modeling and reasoning based solely on abstract conditional independence statements, without a priori commitment to a particular family of graphical models \citep{dawid:21}.

Even then, the structural knowledge expressed by the factorization in Eq.~\eqref{eq:ifm} may be uninformative. 
Depending on the nature of $\Sigma_{\train}$ and $\Sigma_{\test}$, it may be the case that we cannot generalize from training to test environments, and $p(x; \sigma^\star)$ is unidentifiable for all \mbox{$\sigma^\star \in \Sigma_{\test}$}.
However, all methods for causal inference rely on a \mbox{trade-off} between assumptions and informativeness. 
For example, unmeasured confounding may imply no independence constraints, but modeling unmeasured confounding is challenging, even more so for equilibrium data without observable dynamics. 
If we can get away with pure factorization constraints implied by an array of experimental conditions and domain knowledge, 
we should embrace this opportunity. 
This is what is done, for instance, in the literature on causal bandits and causal Bayesian optimization \citep{lattimore:2016, sanghack:2018,aglietti:20,sussex:23}, which leverage similar assumptions to decide what to do next. 
However, we are \emph{not} proposing a method for bandits, Bayesian optimization, or active learning. 
The task of estimating $p(x; \sigma^\star)$ and $\mu(\sigma^\star)$ for a novel regime is relevant in itself.
%

\section{Interventional Factor Model: Identification}
\label{sec:ifm_model}
We define our \emph{identification problem} as follows: 
given the \textit{population} distributions
$\mathbb P(\Sigma_{\train}) := \{p(x; \sigma^1), p(x; \sigma^2), \dots, p(x; \sigma^t)\}$
for the training regimes \mbox{$\sigma^j \in \Sigma_{\train}$}, 
and knowledge of the factorization assumptions as given by Eq.~\eqref{eq:ifm}, 
can we identify a given $p(x; \sigma^\star)$ corresponding to some  unobserved test regime \mbox{$\sigma^\star \in \Sigma_{\test}$}?
\begin{figure}[!t]
  \raisebox{1.89cm}{
  \begin{minipage}{0.4\textwidth}
  \hspace{-0.6in}{\bf a.}\\
  \centering
  \begin{tikzpicture}
    \node[obs] (X) at (1.2, 0) {$X$};
    \node[fact] (F1) at (0.6, 1) {\textcolor{white}{$f_1$}};
    \node[fact] (F2) at (1.8, 1) {\textcolor{white}{$f_2$}};;
    \node[intv] (S1) at (0, 2) {$\sigma_1$};
    \node[intv] (S2) at (1.2, 2) {$\sigma_2$};
    \node[intv] (S3) at (2.4, 2) {$\sigma_3$};
    \edge[black]{S1}{F1};
    \edge[black]{S2}{F1};
    \edge[black]{S2}{F2};
    \edge[black]{S3}{F2};
    \uedge[black]{X}{F1};
    \uedge[black]{X}{F2};
  \end{tikzpicture}%
  \end{minipage}
  \begin{minipage}{0.4\textwidth}
  \hspace{-0.2in}{\bf b.}\\
  \\
  \begin{tikzpicture}
    \node[hobs] (F12) at (0, 3) {$(1, 2)$};
    \node[hintv] (FS2) at (1.2, 3) {$2$};
    \node[hobs] (F23) at (2.4, 3) {$(2, 3)$};
    \uedge[black]{F12}{FS2};
    \uedge[black]{FS2}{F23};
  \end{tikzpicture}%
  \end{minipage}
  \begin{minipage}{0.4\textwidth}
  \hspace{-0.2in}{\bf c.}\\
     \begin{tabular}{ccc}
     $\sigma_1$ & $\sigma_2$ & $\sigma_3$\\
     \hline
      0 & 0 & 0\\
      0 & 1 & 0\\
      1 & 0 & 0\\
      1 & 1 & 0\\
      0 & 0 & 1\\
      0 & 1 & 1\\
     \end{tabular}
 \end{minipage}}\\[-5mm]
 \vspace{-5mm}
 \caption{\textbf{(a)} An interventional factor model (IFM) with three (binary) intervention variables. 
  \textbf{(b)} 
 The junction tree of the \emph{$\sigma$-graph} associated with the IFM in \textbf{(a)}:  
  intervention variables are arranged as a hypergraph, 
  where the hypervertices 
  represent the sets of intervention variables that share a factor 
  and the edge represents the overlap between the two sets of intervention variables. 
 %
 %
 \textbf{(c)} A table displaying the assignments for regimes in $\Sigma_{\train}$. 
  From this $\Sigma_{\train}$ and the assumptions in \textbf{(a)}, it is possible to generalize to the two missing regimes, i.e.,  \mbox{$\sigma_1 = \sigma_2 = \sigma_3 = 1$}, and \mbox{$\sigma_1 = \sigma_3 = 1$, $\sigma_2 = 0$}. 
 }
 \label{fig:identifiability_illustration}
 \vspace{-4mm}
\end{figure}
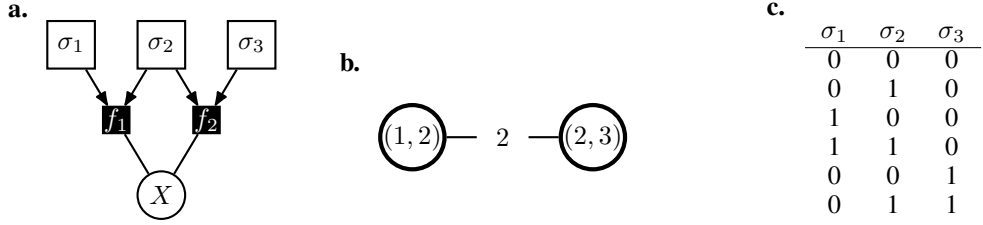

This identification problem is not analyzed in setups such as causal Bayesian optimization \citep{aglietti:20, sussex:23}, which build on DAGs with interventions targeting single variables. 
Without this analysis, it is unclear to what extent the learning 
might be attributed to artifacts due to the choice of prior distribution or regularization, particularly for a sparsely populated $\Sigma_{\train}$. 
As an extreme case that is not uncommon in practice, if we have binary intervention variables and  never observe more than one $\sigma_i$ set to $1$ within the same experiment, it is unclear why we should expect a \mbox{non-linear} model to provide information about pairs of assignments \mbox{$\sigma_i = \sigma_j = 1$} using an \mbox{off-the-shelf} prior or regularizer. 
%
Although eventually a dense enough process of exploration will enrich the database of experiments, this process may be slow and less suitable to situations where the goal is not just to maximize some expected reward but to provide a more extensive picture of the \mbox{dose-response} relationships in a system.
%
%
%
While the smoothness of $p(x; \sigma)$ as a function of $\sigma$ is a relevant \mbox{domain-specific} information, 
it complicates identifiability criteria without further assumptions. 
In what follows, we assume no smoothness conditions, meaning that for some pair \mbox{$p(x; \sigma^a), p(x; \sigma^b)$}, with \mbox{$a \neq b$}, these probability density/mass functions are allowed to be arbitrarily different. 
Such setting is particularly suitable for situations where interventions do take categorical levels, with limited to no magnitude information. 

\paragraph{Identification: preliminaries.} 
Before introducing the main result of this section, let us start by considering the toy example displayed in  Fig.~\ref{fig:identifiability_illustration}: Fig.~\ref{fig:identifiability_illustration}\textbf{(a)} shows the graphical model (how $X$ could be factorized is not relevant here), and Fig.~\ref{fig:identifiability_illustration}\textbf{(c)} shows the assignments for the training regimes $\Sigma_{\train}$ (all intervention variables are binary). 
This training set lacks the experimental assignment \mbox{$\sigma_1=\sigma_2=\sigma_3=1$}.\footnote{As well as the assignment \mbox{$\sigma_1=\sigma_3=1, \sigma_2=0$}, which can be recovered by an analogous argument.} 
However, this regime is implied by the model and $\Sigma_{\train}$. 
To see this, consider the factorization \mbox{$p(x; \sigma) \propto f_1(x; \sigma_1, \sigma_2)f_2(x; \sigma_2, \sigma_3)$}, it implies:  
\[
\frac{p\big(x; (1, 1, 0)\big)}{p\big(x; (0, 1, 0)\big)} \propto \frac{f_1\big(x; (1, 1)\big)}{f_1\big(x; (0, 1)\big)},
\]
\noindent where we used \mbox{$(1, 1, 0)$} etc. to represent the $\sigma$ assignments. 
Because both \mbox{$(1, 1, 0)$} and \mbox{$(0, 1, 0)$} are in $\Sigma_{\train}$, the ratio is identifiable up to a multiplicative constant.
Moreover, multiplying and dividing the result by \mbox{$f_2\big(x; (1, 1)\big)$}, we get:
\[
\frac{p\big(x; (1, 1, 0)\big)}{p\big(x; (0, 1, 0)\big)} \propto \frac{f_1\big(x; (1, 1)\big)}{f_1\big(x; (0, 1)\big)}\frac{f_2\big(x; (1, 1)\big)}{f_2\big(x; (1, 1)\big)} \propto 
\frac{p\big(x; (1, 1, 1)\big)}{p\big(x; (0, 1, 1)\big)},
\]
\noindent from which, given that \mbox{$(0, 1, 1)$} is also in $\Sigma_{\train}$, we can derive \mbox{$p\big(x; (1, 1, 1)\big)$}.

\subsection{Message Passing Formulation}
The steps in this reasoning can be visualized in Fig.~\ref{fig:identifiability_illustration}\textbf{(b)}. 
The leftmost \emph{hypervertex} represents $(\sigma_1, \sigma_2)$ and suggests $\sigma_1$ can be isolated from $\sigma_3$. 
The first ratio $p\big(x; (1, 1, 0)\big) / p\big(x; (0, 1, 0)\big)$ considers three roles: $\sigma_1$ can be isolated (it is set to a ``baseline'' of $0$ in the denominator) and $\sigma_3$ is yet to be considered (it is set to the baseline value of $0$ in the numerator). This ratio sets the stage for the next step where $(\sigma_2, \sigma_3)$ is ``absorbed'' in the construction of the model evaluated at $(1, 1, 1)$. 
The entries in $\Sigma_{\train}$ were chosen so that we see all four combinations of $(\sigma_1, \sigma_2)$ and all four combinations of $(\sigma_2, \sigma_3)$, while avoiding the requirement of seeing all eight combinations of $(\sigma_1, \sigma_2, \sigma_3)$. 
How to generalize this idea is the challenge. Next, we present our first proposed solution. 

\paragraph{Definitions.} 
Before we proceed, we need a few definitions. 
First, recall that we represent the \emph{baseline} (or ``observational'') regime as \mbox{$\sigma_1 = \sigma_2 = \dots = \sigma_d = 0$}.  
This describes data captured under a default protocol, e.g., a transcriptomic study in which no genes are knocked down.
Let \smash{$\Sigma_{[Z]}^0$} denote the set of all environments \mbox{$\sigma^j \in \Sigma$} such that \mbox{$\sigma_i^j = 0$} if \mbox{$i \not \in Z$}. 
For instance, for the example in Fig.~\ref{fig:identifiability_illustration}, 
a set \smash{\mbox{$Z := \{2\}$} implies that \smash{\mbox{$\Sigma_{[Z]}^0 = \big\{(0, 0, 0), (0, 1, 0)\big\}$}}}. 
Finally, let $\sigma^{[Z(\star)]}$ be the intervention vector given by \mbox{$\sigma_i = \sigma^{\star}_i$}, if \mbox{$i \in Z$}, and $0$ otherwise.
%
In what follows, we also use of the concepts of graph \emph{decompositions}, \emph{decomposable graphs} and \emph{junction trees}, as commonly applied to graphical models \citep{lauritzen:96, cowell:99}. 
In Appendix~\ref{appendix:proof}, we review these concepts.\footnote{\textbf{Quick summary}: a junction tree is formed by turning the cliques of an undirected graph into the (hyper)vertices of the tree. Essential to the definition, a junction tree $\mathcal T$ must have a \emph{running intersection property}: given an intersection \mbox{$S := H_i \cap H_j$} of any two hypervertices \mbox{$H_i, H_j$}, all hypervertices in the unique path in $\mathcal T$ between $H_i$ and $H_j$ must contain $S$. This property captures the notion that satisfying local agreements (``equality of properties'' of subsets of elements between two adjacent hypervertices in the tree) should imply global agreements. A decomposable graph is simply a graph whose cliques can be arranged as a junction tree.} 
An IFM $\mathcal I$ with intervention vertices \mbox{$\sigma_1, \dots, \sigma_d$} has an associated \emph{$\sigma$-graph} denoted by \mbox{$\mathcal G_{\sigma(\mathcal I)}$}, defined as an undirected graph with vertices \mbox{$\sigma_1, \dots, \sigma_d$}, and where edge \mbox{$\sigma_i - \sigma_j$} is present if and only if $\sigma_i$ and $\sigma_j$ are simultaneously present in at least one common factor $f_k$ in $\mathcal I$.  
For example, the \mbox{$\sigma$-graph} of the IFM represented in Fig.~\ref{fig:identifiability_illustration}\textbf{(a)} is \mbox{$\sigma_1 - \sigma_2 - \sigma_3$}. 
This is a decomposable graph with vertex partition \mbox{$A = \{\sigma_1\}$, $B = \{\sigma_2\}$ and $C = \{\sigma_3\}$}. 
As this $\sigma$-graph is an undirected decomposable graph, it has a junction tree, which we depict in Fig.~\ref{fig:identifiability_illustration}\textbf{(b)}.

\paragraph{Identification: message passing formulation.} Let $\mathcal I$ be an IFM representing a model for the $m$-dimensional random vector $X$ under the $d$-dimensional intervention vector $\sigma \in \Sigma$. Model $\mathcal I$ has unknown factor parameters but a known factorization $p(x; \sigma) \propto \prod_k f_k(x_{S_k}; \sigma_{F_k})$.

\begin{theorem}
 Assume that the $\sigma$-graph $\mathcal G_{\sigma(\mathcal I)}$ is decomposable. 
 Given a set $\Sigma_{\train} = \{\sigma^1, \dots, \sigma^t\} \subseteq \Sigma$ and known distributions $p(x; \sigma^1), \dots, p(x; \sigma^t)$, the following conditions are sufficient to identify any $p(x; \sigma^\star)$, $\sigma^\star \in \Sigma$:\\[3pt] 
 i) all distributions indexed by $\sigma^1, \dots, \sigma^t, \sigma^\star$ have the same support; and \\[3pt]
 ii) $\Sigma_{[F_k]}^0 \in \Sigma_{\train}$ for all $F_k$ in the factorization of $\mathcal I$.\\[3pt] 
 The algorithm for computing $p(x; \sigma^\star)$ works as follows. Construct a junction tree $\mathcal T$ for $\mathcal I$, choose an arbitrary vertex in $\mathcal T$ to be the root, and direct $\mathcal T$ accordingly. If $V_k$ is a hypervertex in $\mathcal T$, let $D_k$ be the union of all intervention variables contained in the descendants of $V_k$ in $\mathcal T$. Let $B_k$ be the intersection of the invervention variables contained in $V_k$ with the intervention variables contained in the parent $V_{\pi(k)}$ of $V_k$ in $\mathcal T$.\\[3pt]
 We define a \emph{message} from a non-root vertex $V_k$ to its parent $V_{\pi(k)}$ as
\begin{equation}
m_k^x := \frac{p(x; \sigma^{[D_k(\star)]})}{p(x; \sigma^{[B_k(\star)]})},
\label{eq:message_body}
\end{equation}
\vspace{-1pt}
\noindent with the update equation
\begin{equation}
p(x; \sigma^{[D_k(\star)]}) \propto
p(x; \sigma^{[F_k(\star)]})\prod_{V_{k'} \in ch(k)} m_{k'}^x,
\label{eq:update_body}
\end{equation}
where the product over $ch(k)$, the children of $V_k$ in $\mathcal{T}$, is defined to be 1 if $ch(k) = \emptyset$. 
\label{theorem:main}
\end{theorem}

(Proofs are provided in Appendix~\ref{appendix:proof}.) 
In particular, if no factor contains more than one intervention variable, the corresponding \mbox{$\sigma$-graph} will be fully disconnected. This happens, e.g, for an IFM derived from a DAG without hidden variables and where each intervention has one child, as in Fig.~\ref{fig:graph_differences}. 


\subsection{Algebraic Formulation}
If a $\sigma$-graph is not decomposable, the usual trick of triangulation prior to clique extraction can be used \citep{lauritzen:96,cowell:99}, at the cost of creating cliques which are larger than the original factors. 
Alternatively, and a generalization of Eqs. (\ref{eq:message_body}) and (\ref{eq:update_body}), we consider transformations of the distributions in $\mathbb P(\Sigma_{\train})$ by products and ratios. We define a \emph{PR-transformation} of a density set \mbox{$\{p(x; \sigma^1), \dots, p(x; \sigma^t)\}$} as any formula \smash{\mbox{$\prod_{i = 1}^t p(x; \sigma^i)^{q_i}$}}, for a collection \mbox{$q_1, \dots, q_t$} of real numbers.

\begin{theorem}
Let $\sigma_{F_k}^v$ denote a particular value of $\sigma_{F_k}$, and let $\mathbb{D}_k$ be the domain of $\sigma_{F_k}$. 
Given a collection $\mathbb P(\Sigma_{\train}) := \{p(x; \sigma^1), \dots, p(x; \sigma^t)\}$ and a postulated model factorization $p(x; \sigma) \propto \prod_{k = 1}^l f_k(x_{S_k}; \sigma_{F_k})$, a sufficient and almost-everywhere necessary condition for a given $p(x; \sigma^\star)$ to be identifiable by PR-transformations of $\mathbb P(\Sigma_{\train})$ is that there exists some solution to the system
\begin{equation}
\displaystyle
\forall k \in \{1, 2, \dots, l\}, \forall \sigma_{F_k}^v \in \mathbb D_k, \left(\sum_{i = 1: \sigma_{F_k}^i = \sigma_{F_k}^v}^t q_i\right) = \mathbbm{1}(\sigma_{F_k}^\star = \sigma_{F_k}^v),
\label{eq:algebraic}
\end{equation}
\noindent where $\mathbbm{1}(\cdot)$ is the indicator function returning 1 or 0 if its argument is true or false, respectively. 
\label{theorem:algebraic}
\end{theorem}

\begin{figure}[!t]
\centering
  \raisebox{1.89cm}{
     \begin{tabular}{c|ccc|cccc|cccc|cccc}
     & $\sigma_1$ & $\sigma_2$ & $\sigma_3$ &
     $f_1^{00}$ & $f_1^{01}$ & $f_1^{10}$ & $f_1^{11}$ &
     $f_2^{00}$ & $f_2^{01}$ & $f_2^{10}$ & $f_2^{11}$ &
     $f_3^{00}$ & $f_3^{01}$ & $f_3^{10}$ & $f_3^{11}$\\
     \hline
      $q_1$ & 0 & 0 & 0 & \checkmark & & & & \checkmark & & & & \checkmark & & &\\
      $q_2$ & 0 & 1 & 0 & & \checkmark & & & & & \checkmark & & \checkmark & & &\\
      $q_3$ & 1 & 0 & 0 & & & \checkmark & & \checkmark & & & & & & \checkmark &\\
      $q_4$ & 1 & 1 & 0 & & & & \checkmark & & & \checkmark& & & & \checkmark &\\
      $q_5$ & 0 & 0 & 1 & \checkmark & & & & & \checkmark & & & & \checkmark & &\\
      $q_6$ & 0 & 1 & 1 & & \checkmark & & & & & & \checkmark & & \checkmark & &\\
      $q_7$ & 1 & 0 & 1 & & & \checkmark & & & \checkmark & & & & & & \checkmark\\
      \hline
      $p^\star$ & 1 & 1 & 1 & 0 & 0 & 0 & 1 & 0 & 0 & 0 & 1 & 0 & 0 & 0 & 1\\
     \end{tabular}
}
\vspace{-7mm}
 \caption{Example of how to infer the target distribution \mbox{$p\big(x; (1, 1, 1)\big)$} from the other seven possible regimes given the postulated factorization 
 \mbox{$p(x; \sigma) \propto f_1(x; \sigma_1, \sigma_2)f_2(x; \sigma_2, \sigma_3)f_3(x; \sigma_1, \sigma_3)$}.
 We use $f_k^{ab}$ as a shorthand notation for \mbox{$f_k(x; \sigma_{k_1} = a, \sigma_{k_2} = b)$}, 
 e.g, \mbox{$p\big(x; (1, 1, 1)\big) \propto f_1^{11}f_2^{11}f_3^{11}$}.
 For a PR-transformation 
 \mbox{$(f_1^{00}f_2^{00}f_3^{00})^{q_1} \times (f_1^{01}f_2^{10}f_3^{00})^{q_2} \times \dots \times (f_1^{10}f_2^{01}f_3^{11})^{q_7}$} 
\mbox{$f_1^{11}f_2^{11}f_3^{11}$}
 up to a multiplicative constant independent of $x$, 
 we need to find a solution satisfying: 
 \mbox{$q_1 + q_5 = 0$}, \mbox{$q_2 + q_6 = 0$}, $\dots$, \mbox{$q_7 = 1$}.    
  (In the table, this can be seen by going through each $f_k^{ab}$ column and adding up the corresponding exponents $q_j$ when there is a tick in row $j$; the sum must agree with the corresponding entry in the final row). 
 A solution is \smash{\mbox{$q_1 = q_4 = q_6 = q_7 = 1$, $q_2 = q_3 = q_5 = -1$}}, corresponding to \smash{\mbox{$p(x; \sigma^\star) \propto \big( p(x; \sigma^1)p(x; \sigma^4)p(x; \sigma^6)p(x; \sigma^7) \big)/ \big( p(x; \sigma^2)p(x; \sigma^3)p(x; \sigma^5)\big)$}}.
 }
 \label{fig:algebraic_method}
 \vspace{-2mm}
\end{figure}

The solution to the system gives the PR-transformation. An example is shown in Fig.~\ref{fig:algebraic_method}. 
The main idea is that, for a fixed $x$, any factor $f_k(x_{S_k}; \sigma_{F_k})$ can algebraically be interpreted as an arbitrary symbol indexed by $\sigma_{F_k}$, say ``$f_k^{\sigma_{F_k}}$''. 
This means that any density $p(x; \sigma^i)$ is (proportional to) a ``squarefree'' monomial $m^i$ in those symbols (i.e, products with exponents 0 or 1). We need to manipulate a set of monomials (the unnormalized densities in $\mathbb P(\Sigma_{\train})$) into another monomial $m^\star$ (the unnormalized $p(x; \sigma^\star)$). 
More generally, the possible set functions $g(\cdot), h(\cdot)$ such that $g(m^\star, \{m^1, \dots, m^t\}) = h(\{m^1, \dots, m^t\})$, with $g(\cdot)$ invertible in $m^\star$, suggest that $g(\cdot)$ and $h(\cdot)$ must be the product function (monomials are closed under products, but not other analytical manipulations), although we stop short of stating formally the conditions required for PR-transformations to be complete in the space of all possible functions of $\mathbb P(\Sigma_{\train})$. 
This would be a stronger claim than Theorem \ref{theorem:algebraic}, which shows that Eq. (\ref{eq:algebraic}) is almost-everywhere complete in the space of PR transformations.

The message passing scheme and the algebraic method provide complementary views, with the former giving a \mbox{divide-and-conquer} perspective that identifies subsystems that can be estimated without requiring changes from the baseline treatment everywhere else. 
The algebraic method is more general, but suggests no hierarchy of simpler problems.
%
These results show that the factorization over $X$ is unimportant for identifiability, which may be surprising. Appendix~\ref{appendix:testability} discusses the consequences of this finding, along with a discussion about requirements on the size of $\Sigma_{\train}$.
\vspace{-5pt}

\section{Learning Algorithms}
\label{sec:learning}
The identification results give a license to choose any estimation method we want if identification is established --- while we must make the assumption of the model factorizing according to a postulated IFM, we are not required to explicitly use a likelihood function. 
Theorems \ref{theorem:main} and \ref{theorem:algebraic} provide ways of constructing a target distribution $p(x; \sigma^\star)$ from products of ratios of densities from \smash{$\mathbb P(\Sigma_{\train})$}. 
This suggests a \mbox{plug-in} approach for estimation: estimate products of ratios using density ratio estimation methods and multiply results. 
However, in practice, we found that fitting a likelihood function directly often works better than estimating the product of density ratios, even for intractable likelihoods. 
We now discuss three strategies for estimating some \smash{\mbox{$\mathbb E[Y; \sigma^\star]$}} of interest. 


\paragraph{Deep \mbox{energy-based} models and direct regression.} 
The most direct learning algorithm is to first maximize the sum of \mbox{log-likelihoods} \smash{\mbox{$\mathcal L(\theta; \mathcal D^1, \dots, \mathcal D^t) := \sum_{i = 1}^t \sum_{j = 1}^{n_i} \log p_{\theta}(x^{i(j)}; \sigma^i)$}}, where $n_i$ is the number of samples in the $i$th regime. 
The parameterization of the model is indexed by a vector $\theta$, which defines $p(\cdot)$ as \smash{\mbox{$\log p_\theta(x; \sigma) := \sum_{k = 1}^l \phi_{\theta_{k, \sigma_{F_k}}}(x_{S_k}) + \text{constant}$}}.
Here, \mbox{$\phi_{\theta}(\cdot)$} is a differentiable \mbox{black-box} function, which in our experiments is a MLP (aka a multilayer perceptron, or feedforward neural network). 
Parameter vector \mbox{$\theta_{k, \sigma_{F_k}}$} is the collection of weights and biases of the MLP, a different instance for each factor $k$ \emph{and} combination of values in $\sigma_{F_k}$. 
In principle, making the parameters smooth functions of $\sigma_{F_k}$ is possible, but in the interest of simplifying the presentation, we instead use a \mbox{look-up} table for completely independent parameters as indexed by the possible values of $\sigma_{F_k}$. 
Parameter set $\theta$ is the union of all \mbox{$\sum_{k = 1}^l \prod_{j \in F_k}|\aleph_j|$} MLP parameter sets.
As maximizing \mbox{log-likelihood} is generally intractable, in our implementation we apply \mbox{pseudo-likelihood} with discretization of each variable $X$ in a grid. 
The level of discretization does not affect the number of parameters, as we take their numerical value as is, renormalizing over the \mbox{pre-defined} grid. 
Score matching \citep{aapo:05}, noise contrastive estimation \citep{gutmann:10} or other variants (e.g., \citep{song:19}) could be used; our pipeline is agnostic to this choice, with further details in Appendix \ref{appendix:learning}. 
We then estimate \mbox{$f_y(x) := \mathbb E[Y \mid x]$} using an \mbox{off-the-shelf} method, which in our case is another MLP. 
For a given $\sigma^\star$, we sample from the corresponding $p(x; \sigma^\star)$ with Gibbs sampling and average the results of the regression estimate \smash{$\widehat f_y(x)$} to obtain an estimate \smash{$\widehat{\mu}(\sigma^\star)$}. 

\paragraph{Inverse probability weighting (IPW).} 
A more direct method is to reweight each training sample by the target distribution $p(x; \sigma^\star)$ to generate 
\mbox{$\widehat \mu(\sigma^{i\star}) := \sum_{j = 1}^{n_i} y^{i(j)}w^{i(j)^\star}$}, where $w^{i(j)^\star}$ is the density ratio \mbox{$p(x^{i(j)}; \sigma^\star) / p(x^{i(j)}; \sigma^i)$}, rescaled such that \mbox{$\sum_{j = 1}^{n_i}w^{i(j)^\star} = 1$}. 
There are several direct methods for density ratio estimation \citep{lu2023rethinking} that could be combined using the messages/product-ratios of the previous section, but we found that it was stabler to just take the density ratio of the fitted models using deep \mbox{energy-based} learning, just like in the previous algorithm. 
Once estimators \mbox{$\widehat \mu(\sigma^{\star(1)}), \dots, \widehat \mu(\sigma^{\star(t)})$} are obtained, we combine them by the usual inverse variance weighting rule, \mbox{$\widehat \mu(\sigma^{i\star}) := \sum_{i = 1}^t r^i \mu_{\sigma^{\star(i)}} / \sum_{i = 1}^t r^i$}, where \mbox{$r^i := 1 / \widehat v^{i}$}, and \mbox{$\widehat v^i := \sum_{j = 1}^{n_i} (y^{i(j)})^2 (w^{i(j)^\star})^2$}. 
This method requires neither a model for $f_y(x)$ nor Markov chain Monte Carlo. 
However, it may behave more erratically than the direct method described above, particularly under strong shifts in distribution.

\paragraph{Covariate shift regression.} 
Finally, a third approach for estimating $\mu(\sigma^\star)$ is to combine models for $f_y(x)$ with density ratios, learning a customized $\widehat f_y(x)$ for each test regime separately. A
s this is very slow and did not appear to be advantageous compared to the direct method, we defer a more complete description to Appendix \ref{appendix:covariate_shift}.

\paragraph{Predictive Coverage.}
Even when treatment effects are identifiable within the IFM framework, the uncertainty of resulting estimates can vary widely depending on the training data and learning algorithm. Building on recent work in conformal inference \citep{vovk2005, Lei2018, tibshirani2019}, we derive the following finite sample coverage guarantee for potential outcomes, which requires no extra assumptions beyond those stated above. 

\begin{theorem}[Predictive Coverage.]\label{thm:conform}
    Assume the identifiability conditions of Theorems \ref{theorem:main} or \ref{theorem:algebraic} hold. 
    Fix a target level \mbox{$\alpha \in (0,1)$}, and let $\mathcal{I}_1, \mathcal{I}_2$ be a random partition of observed regimes intro training and test sets of size $n/2$. 
    Fit a model $\widehat{\mu}$ using data from $\mathcal{I}_1$ and compute conformity scores \mbox{$s^{(i)} = |y^{k(i)} - \widehat{\mu}(\sigma^k)|$} using data from $\mathcal{I}_2$. For some new test environment $\sigma^\star$, compute the normalized likelihood ratio \mbox{$w^{(i)}(\sigma^\star) \propto p(x^{k(i)}; \sigma^\star) / p(x^{k(i)}; \sigma^k)$}, rescaled to sum to $n/2$.
    Let $\widehat{\tau}(\sigma^\star)$ be the $q^{\text{th}}$ smallest value in the reweighted empirical distribution $\sum_i w^{(i)}(\sigma^\star) \cdot \delta(s^{(i)})$, where $\delta$ denotes the Dirac delta function and \mbox{$q = \lceil (n/2 + 1) (1 - \alpha) \rceil$}.\\
    Then for any new sample $n+1$, we have: 
    \begin{align*}
        \mathbb{P}\big(Y^{\star(n+1)} \in \widehat{\mu}(\sigma^\star) \pm \widehat{\tau}(\sigma^*) \big) \geq 1 - \alpha.
    \end{align*}
    Moreover, if weighted conformity scores have a continuous joint distribution, then the upper bound on this probability is \mbox{$1 - \alpha + 1/(n/2+1)$}. 
\end{theorem}


\vspace{-3pt}
\section{Experiments}
\label{sec:experiments}
We run a number of \mbox{semi-synthetic} experiments to evaluate the performance of the IFM approach on a range of intervention generalization tasks. 
In this section, we summarize our experimental \mbox{set-up} and main results. 
The code for reproducing all results and figures is available online\footnote{\url{https://github.com/rbas-ucl/intgen}};  
in Appendix~\ref{appendix:experimental_details}, we provide a detailed description of the datasets and models; and in Appendix~\ref{appendix:further_results} we present further analysis and results. 
%

\paragraph{Datasets.} 
Our experiments are based on the following two biomolecular datasets:  
    \textit{i)} \textbf{Sachs} \cite{sachs2005causal}:  
    a cellular signaling network with $11$ nodes representing phosphorylated proteins and phospholipids, 
    several of which were perturbed with targeted reagents to stimulate or inhibit expression. 
    There are $4$ binary intervention variables. 
        $\Sigma_{\train}$ has $5$ regimes: a \emph{baseline}, with all \mbox{$\sigma_i = 0$}, and $4$ ``\mbox{single-intervention}'' regimes, each with a different single \mbox{$\sigma_i = 1$}. 
    $\Sigma_{\test}$ consist of the remaining $11$ (\mbox{$= 2^4 - 5$}) unseen regimes. 
    \textit{ii)} \textbf{DREAM} \cite{greenfield2010dream4}: 
    Simulated data based on a known \textit{E.~coli} regulatory \mbox{sub-network} with $10$ nodes.
    There are $10$ binary intervention variables, 
    and (similarly) $\Sigma_{\train}$ has a baseline regime, with all \mbox{$\sigma_i = 0$}, and $10$ \mbox{single-intervention} regimes, with single choices of \mbox{$\sigma_i = 1$}. 
    $\Sigma_{\test}$ consists of $45$ (\mbox{$10 \times 9 / 2=45$}) unseen regimes defined by all pairs \mbox{$\sigma_i = \sigma_j = 1$}.

\paragraph{Oracular simulators.}
The first step to test intervention generalization is to build a set of proper test beds (i.e., simulators that serve as causal effect oracles for any \mbox{$\sigma \in \Sigma$}), motivated by expert knowledge about the underlying system dynamics. 
Neither the original Sachs data nor the DREAM simulator we used provide joint intervention data required in our evaluation. 
Thus, for each domain, we trained two ground truth simulators: 
\textit{i)} \textbf{Causal-DAG}: a DAG model following the DAG structure and data provided by the original Sachs et al. 
and DREAM sources
(DAGs shown in Fig.~\ref{fig:graph_sachs}(\textbf{a})) and Fig.~\ref{fig:graph_dream}(\textbf{a}), respectively). 
Given the DAG, we fit a model where each conditional distribution is a heteroskedastic Gaussian with mean and variance parameterized by MLPs
(with $10$ hidden units) 
of the respective parents. 
\textit{ii)} \textbf{Causal-IFMs}: 
the corresponding IFM is obtain by a direct projection of the postulated DAG factors (as done in, e.g., Fig.~\ref{fig:graph_differences}\textbf{(b)}). 
The likelihood is a neural energy model (Section~\ref{sec:learning}) with MLPs 
with $15$ hidden units defining potential functions. 
After fitting these models, we compute by Monte Carlo simulation their implied ground truths for every choice of $\sigma^{*}$. 
The outcomes $Y$ are then 
generated under $100$ different structural equations of the form \smash{\mbox{$\tanh(\lambda^\top X) + \epsilon$}}, with random independent normal weights $\lambda$ and \smash{\mbox{$\epsilon \sim \mathcal{N}(0, v_y)$}}. 
$\lambda$ and $v_y$ are scaled such that the ground truth variance of \smash{\mbox{$\lambda^{\mathsf T}X$}}, \smash{\mbox{$\text{Var}(\lambda^{\mathsf T}X)$}}, is sampled uniformly at random from the interval \mbox{$[0.6, 0.8]$}, and set \smash{\mbox{$v_y := 1 - \text{Var}(\lambda^{\mathsf T}X)$}}. 

\paragraph{Compared models.} We implement three variants of our proposed IFM model, corresponding to the three learning algorithms described in Section \ref{sec:learning}: \textit{i)} \textbf{IFM1} uses deep \mbox{energy-based} models and direct regression; \textit{ii)} \textbf{IFM2} uses an IPW estimator; and \textit{iii)} \textbf{IFM3} relies on covariate shift regression. 
We compare these models to the following benchmarks: 
\textit{i)} \textbf{Black-box}: We apply an \mbox{off-the-shelf} algorithm (XGBoost \citep{chen_xgboost}) to learn a direct mapping from $\sigma$ to $Y$ without using  $X$. 
This model does not exploit any structural assumptions.
\textit{ii)} \textbf{DAG}: We estimate the structural equations in an acyclic topological ordering that is consistent with the data generating process. 
The likelihood is defined by conditional Gaussian models with mean and variances parameterized as MLPs, matching exactly one of the simulators described below. 
This gives this competitor much advantage in the benchmarks generated by DAGs, as following a parametric Gaussian with additive error structure is substantive information to be exploited.

\begin{figure}[t]
  \centering
  \includegraphics[width=0.99\textwidth]{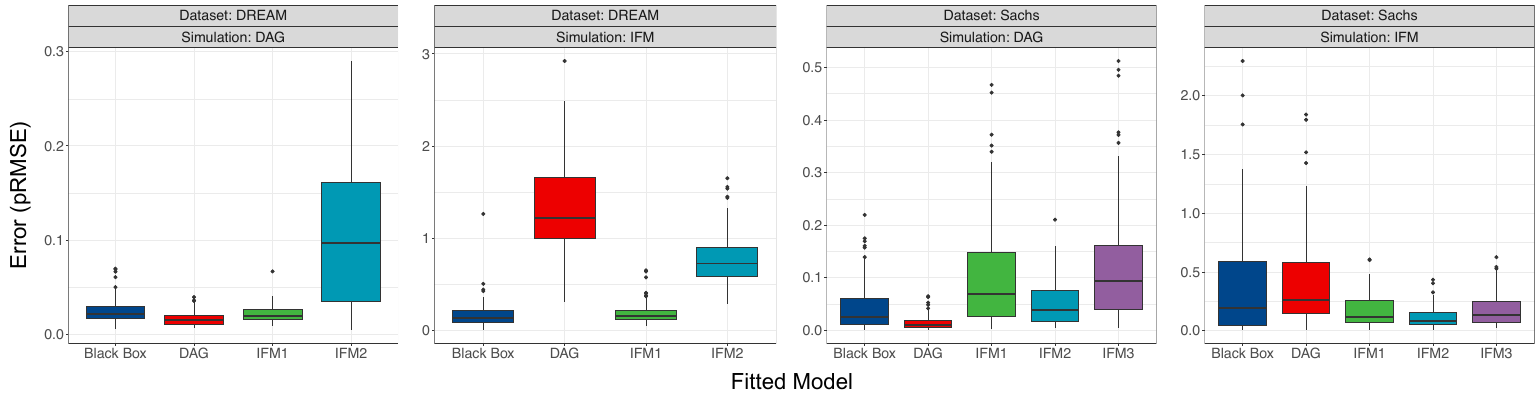}
  \vspace{-6pt}
  \caption{Experimental results on a range of intervention generalization tasks, see text for details.}
  \label{fig:boxes}
\end{figure}

\paragraph{Results.}
We evaluate model performance based on the proportional root mean squared error (pRMSE), defined as the average of the squared difference between the ground truth \smash{$Y$} and estimated \smash{$\widehat{Y}$}, with each entry further divided by the ground truth variance of the corresponding \smash{$Y$}. 
Results are visualized in Fig.~\ref{fig:boxes}. 
We additionally run a series of \mbox{one-sided} binomial tests to determine whether models significantly outperform the black box baseline, and compare the Spearman's rank correlation \cite{zar2014spearman} between expected and observed outcomes for all test regimes. 
Unsurprisingly, DAGs do best when the ground truth is a \mbox{Causal-DAG}, while IFM methods do better when the data generator is a \mbox{Causal-IFM}. 
%
Still, some IFMs are robust to both ground truth models, with IFM1 (the deep energy model) doing especially well on the DREAM dataset, and IFM2 (the IPW estimator) excelling on the Sachs data. 

In particular, for the DREAM datasets, IFM1 (scaled) errors remain stable, while the DAG has a major loss of performance when a \mbox{non-DAG} ground truth is presented.  
However, IFM2 underperform on DREAM datasets, which might be because the knockdowns regimes in those datasets induce dramatic shifts in distribution overlap among regimes. 
(We omit results for IFM3 (covariate shift regression) on these datasets, as the method does not converge in a reasonable time.)
%
Though the \mbox{black-box} method (XGBoost algorithm and no structural assumption) sometimes struggles to extrapolate, displaying a \mbox{long-tailed} distribution of errors, 
it actually does surprisingly well in some experiments, especially on the DREAM dataset. 
In fact, the \mbox{black-box} method is essentially indistinguishable from linear regression in this benchmark (results not shown), which is not surprising given the sparsity of the training data.
However, the \mbox{non-additivity} of the $\sigma$ effect on $Y$ is more prominent in the Sachs datasets, making it difficult to generalize without structural assumptions. 
%
%


\vspace{-7pt}

\section{Related Work}
\label{sec:related}
A factor graph interpretation of Pearl's \mbox{\emph{do}-calculus} framework is described in \citep{winn:12}, but without addressing the problem of identifiability. 
More generally, several authors have exploit structural assumptions to predict the effects of unseen treatments.  
Much of this research falls under the framework of \textit{transportability} \citep{Pearl_Bareinboim_og, bareinboim_transportability}, where the goal is to identify causal estimands from a combination of observational and experimental data collected under different regimes. 
If only atomic interventions are considered, the \mbox{$do$-calculus} is sound and complete for this task \citep{Lee_generalized}. 
The \mbox{$\sigma$-calculus} introduced by \citep{forre2018constraint, correa_sigma_calc} extends these transportability results to soft interventions \citep{correa_completeness_sigma, correa_counterfactuals}. 
%
However, these methods are less clearly defined when an intervention affects several variables simultaneously,  
and while DAGs with additive errors have some identifiability \citep{saengkyongam:20} and estimation \citep{sussex:23} advantages, acyclic error additivity may not be an appropriate
assumption in some domains. 
Under further parametric assumptions, causal effects can be imputed with matrix completion techniques \citep{squires2022causal, agarwal_ab} or more generic supervised learning approaches \citep{gultchin_complex}, but these methods often require some data to be collected for all regimes in $\Sigma_{\test}$. 
Finally, \citep{agarwal_combo} is the closest related work in terms of goals, 
factorizing the function space of each $\mathbb E[X_i; \sigma]$ directly as a function of $\sigma$.
%

Another strand of related research pertains to online learning settings. 
For instance, several works have shown that causal information can boost convergence rates in \mbox{multi-armed} bandit problems when dependencies are present between arms \citep{lattimore:2016, sanghack:2018, dekroon:2020}, even when these structures must themselves be adaptively learned \cite{bilodeau2022adaptively}. 
This suggests a promising direction for future work, where identification strategies based on the IFM framework are used to prioritize the search for optimal treatments.
Indeed, combining samples across multiple regimes can be an effective strategy for causal discovery, as illustrated by recent advances in invariant causal prediction \citep{peters2016, Heinze2018, pfister2019, zhang2020}, 
and it can also help with domain adaptation and covariate shift \citep{magliacane_da, arjovsky2020invariant, buhlmann2020, chen_jmlr}. 
\vspace{-7pt}


\section{Conclusion}
We introduced the IFM framework for solving intervention generalization tasks. 
Results from simulations calibrated by \mbox{real-world} data show that our method successfully predicts outcomes for novel treatments, providing practitioners with new methods for conducting synthetic experiments. 
Future work includes: 
\emph{i)} integrating the approach with experimental design, Bayesian optimization and bandit learning; 
\emph{ii)} variations that include \mbox{pre-treatment} variables and generalizations across heterogeneous subpopulations, inspired by complementary matrix factorization methods such as \citep{agarwal_combo}; 
\emph{iii)} tackling sequential treatments; and 
\emph{iv)} diagnostics of \mbox{cross-regime} overlap issues \citep{hill:11, oberst:20}.






\newpage

\begin{ack}
We thank the anonymous reviewers and Mathias Drton for useful discussions. GBH was supported by the ONR grant 62909-19-1-2096, and JY was supported by the EPSRC grant EP/W024330/1. RS was partially supported by both grants. JZ was supported by UKRI grant EP/S021566/1.
\end{ack}

{\small
\bibliography{biblio,rbas}

\begin{thebibliography}{}

\bibitem[Abbeel et~al., 2006]{abbeel:06}
Abbeel, P., Koller, D., and Ng, A.~Y. (2006).
\newblock Learning factor graphs in polynomial time and sample complexity.
\newblock {\em Journal of Machine Learning Research}, 7:1743--1788.

\bibitem[Agarwal et~al., 2023]{agarwal_combo}
Agarwal, A., Agarwal, A., and Vijaykumar, S. (2023).
\newblock Synthetic combinations: A causal inference framework for
  combinatorial interventions.
\newblock \emph{arXiv} preprint, 2303.14226.

\bibitem[Agarwal et~al., 2020]{agarwal_ab}
Agarwal, A., Shah, D., and Shen, D. (2020).
\newblock Synthetic {A/B} testing using synthetic interventions.
\newblock \emph{arXiv} preprint, 2006.07691.

\bibitem[Aglietti et~al., 2020]{aglietti:20}
Aglietti, V., Lu, X., Paleyes, A., and Gonz\'{a}lez, J. (2020).
\newblock Causal {B}ayesian optimization.
\newblock {\em $23^{rd}$ International Conference on Artificial Intelligence
  and Statistics}.

\bibitem[Akbari et~al., 2023]{akbari:23}
Akbari, K., Winter, S., and Tomko, M. (2023).
\newblock Spatial causality: A systematic review on spatial causal inference.
\newblock {\em Geographical Analysis}, 55:56--89.

\bibitem[Arjovsky et~al., 2020]{arjovsky2020invariant}
Arjovsky, M., Bottou, L., Gulrajani, I., and Lopez-Paz, D. (2020).
\newblock Invariant risk minimization.
\newblock \emph{arXiv} preprint, 1907.02893.

\bibitem[Bareinboim and Pearl, 2014]{bareinboim_transportability}
Bareinboim, E. and Pearl, J. (2014).
\newblock Transportability from multiple environments with limited experiments:
  Completeness results.
\newblock In {\em Advances in Neural Information Processing Systems},
  volume~27.

\bibitem[Bica et~al., 2021]{bica2021}
Bica, I., Alaa, A.~M., Lambert, C., and van~der Schaar, M. (2021).
\newblock From real-world patient data to individualized treatment effects
  using machine learning: Current and future methods to address underlying
  challenges.
\newblock {\em Clinical Pharmacology \& Therapeutics}, 109(1):87--100.

\bibitem[Bilodeau et~al., 2022]{bilodeau2022adaptively}
Bilodeau, B., Wang, L., and Roy, D.~M. (2022).
\newblock Adaptively exploiting \textit{d}-separators with causal bandits.
\newblock In {\em Advances in Neural Information Processing Systems}.

\bibitem[Bishop et~al., 1974]{bishop:74}
Bishop, Y. M.~M., Fienberg, S.~E., and Holland, P.~W. (1974).
\newblock {\em Discrete Multivariate Analysis: Theory and Practice}.
\newblock MIT Press.

\bibitem[Blom and Mooij, 2023]{blom:23}
Blom, T. and Mooij, J.~M. (2023).
\newblock Causality and independence in perfectly adapted dynamical systems.
\newblock {\em Journal of Causal Inference}, 11(1):2885--2915.

\bibitem[Bongers et~al., 2021]{bongers:21}
Bongers, S., Forr\'{e}, P., Peters, J., and Mooij, J.~M. (2021).
\newblock Foundations of structural causal models with cycles and latent
  variables.
\newblock {\em Annals of Statistics}, 49(5):2885--2915.

\bibitem[B{\"u}hlmann, 2020]{buhlmann2020}
B{\"u}hlmann, P. (2020).
\newblock {Invariance, Causality and Robustness}.
\newblock {\em Statistical Science}, 35(3):404 -- 426.

\bibitem[Chen and Guestrin, 2016]{chen_xgboost}
Chen, T. and Guestrin, C. (2016).
\newblock {XGB}oost: A scalable tree boosting system.
\newblock In {\em Proceedings of the 22nd ACM SIGKDD International Conference
  on Knowledge Discovery and Data Mining}, page 785–794.

\bibitem[Chen and B\"{u}hlmann, 2021]{chen_jmlr}
Chen, Y. and B\"{u}hlmann, P. (2021).
\newblock Domain adaptation under structural causal models.
\newblock {\em J. Mach. Learn. Res.}, 22(1).

\bibitem[Constantinou and Dawid, 2017]{constantinou:17}
Constantinou, P. and Dawid, A.~P. (2017).
\newblock Extended conditional independence and applications in causal
  inference.
\newblock {\em The Annals of Statistics}, 45(6):2618--2653.

\bibitem[Correa and Bareinboim, 2020a]{correa_sigma_calc}
Correa, J. and Bareinboim, E. (2020a).
\newblock A calculus for stochastic interventions:causal effect identification
  and surrogate experiments.
\newblock {\em Proceedings of the AAAI Conference on Artificial Intelligence},
  34(06):10093--10100.

\bibitem[Correa and Bareinboim, 2020b]{correa_completeness_sigma}
Correa, J. and Bareinboim, E. (2020b).
\newblock General transportability of soft interventions: Completeness results.
\newblock In {\em Advances in Neural Information Processing Systems},
  volume~33, pages 10902--10912.

\bibitem[Correa et~al., 2022]{correa_counterfactuals}
Correa, J., Lee, S., and Bareinboim, E. (2022).
\newblock Counterfactual transportability: A formal approach.
\newblock In {\em International Conference on Machine Learning}.

\bibitem[Cowell et~al., 1999]{cowell:99}
Cowell, R., Dawid, A., Lauritzen, S., and Spiegelhalter, D. (1999).
\newblock {\em Probabilistic Networks and Expert Systems}.
\newblock Springer-Verlag.

\bibitem[Dash, 2005]{dash:05}
Dash, D. (2005).
\newblock Restructuring dynamic causal systems in equilibrium.
\newblock {\em Proceedings of the Tenth International Workshop on Artificial
  Intelligence and Statistics}, pages 81--88.

\bibitem[Dawid, 2010]{dawid:10}
Dawid, A.~P. (2010).
\newblock Beware of the {DAG}!
\newblock {\em Proceedings of Workshop on Causality: Objectives and Assessment
  at NIPS 2008, PMLR}, 6:59--86.

\bibitem[Dawid, 2021]{dawid:21}
Dawid, P. (2021).
\newblock Decision-theoretic foundations for statistical causality.
\newblock {\em Journal of Causal Inference}, 9(56):39--77.

\bibitem[de~Kroon et~al., 2020]{dekroon:2020}
de~Kroon, A. A. W.~M., Belgrave, D., and Mooij, J.~M. (2020).
\newblock Causal discovery for causal bandits utilizing separating sets.
\newblock \emph{arXiv}:2009.07916.

\bibitem[Drton, 2009]{drton:09}
Drton, M. (2009).
\newblock Discrete chain graph models.
\newblock {\em Bernoulli}, 15:736--753.

\bibitem[Eaton and Murphy, 2007]{eaton:07}
Eaton, D. and Murphy, K. (2007).
\newblock Exact {B}ayesian structure learning from uncertain interventions.
\newblock {\em Artificial Intelligence \& Statistics}.

\bibitem[Eberhardt and Scheines, 2007]{eberhardt:07}
Eberhardt, F. and Scheines, R. (2007).
\newblock Interventions and causal inference.
\newblock {\em Philosophy of Science}, 74:981--989.

\bibitem[Forr{\'e} and Mooij, 2018]{forre2018constraint}
Forr{\'e}, P. and Mooij, J.~M. (2018).
\newblock Constraint-based causal discovery for non-linear structural causal
  models with cycles and latent confounders.
\newblock In {\em Proceedings of the 34th Annual Conference on Uncertainty in
  Artificial Intelligence}, pages 269--278.

\bibitem[Gentzel et~al., 2021]{gentzel2021and}
Gentzel, A.~M., Pruthi, P., and Jensen, D. (2021).
\newblock How and why to use experimental data to evaluate methods for
  observational causal inference.
\newblock In {\em International Conference on Machine Learning}, pages
  3660--3671. PMLR.

\bibitem[Greenfield et~al., 2010]{greenfield2010dream4}
Greenfield, A., Madar, A., Ostrer, H., and Bonneau, R. (2010).
\newblock Dream4: Combining genetic and dynamic information to identify
  biological networks and dynamical models.
\newblock {\em PloS one}, 5(10):e13397.

\bibitem[Gultchin et~al., 2021]{gultchin_complex}
Gultchin, L., Watson, D., Kusner, M., and Silva, R. (2021).
\newblock Operationalizing complex causes: A pragmatic view of mediation.
\newblock In {\em Proceedings of the 38th International Conference on Machine
  Learning}, volume 139 of {\em Proceedings of Machine Learning Research},
  pages 3875--3885. PMLR.

\bibitem[Gutmann and Hyv\"{a}rinen, 2010]{gutmann:10}
Gutmann, M. and Hyv\"{a}rinen, A. (2010).
\newblock Noise-contrastive estimation: A new estimation principle for
  unnormalized statistical models.
\newblock {\em 13th International Conference on Artificial Intelligence and
  Statistics (AISTATS)}, pages 297--304.

\bibitem[Heinze-Deml et~al., 2018]{Heinze2018}
Heinze-Deml, C., Peters, J., and Meinshausen, N. (2018).
\newblock {Invariant Causal Prediction for Nonlinear Models}.
\newblock {\em J. Causal Inference}, 6(2).

\bibitem[Hern\'{a}n and J, 2020]{hernan:20}
Hern\'{a}n, M. and J, R. (2020).
\newblock {\em Causal Inference: What If}.
\newblock Chapman $\&$ Hall/CRC.

\bibitem[Higbee, 2023]{higbee:23}
Higbee, S. (2023).
\newblock Policy learning with new treatments.
\newblock {\em arXiv:2210.04703}.

\bibitem[Hill and Su, 2011]{hill:11}
Hill, J. and Su, Y. (2011).
\newblock Assessing lack of common support in causal inference using bayesian
  nonparametrics: Implications for evaluating the effect of breastfeeding on
  children’s cognitive outcomes.
\newblock {\em The Annals of Applied Statistics}, 7:1386--1420.

\bibitem[Hoover, 2001]{hoover:01}
Hoover, K. (2001).
\newblock {\em Causality in Macroeconomics}.
\newblock Cambridge University Press.

\bibitem[Hyv\"{a}rinen, 2005]{aapo:05}
Hyv\"{a}rinen, A. (2005).
\newblock Estimation of non-normalized statistical models by score matching.
\newblock {\em Journal of Machine Learning Research}, 6:695--709.

\bibitem[Hyv\"{a}rinen et~al., 2008]{aapo:08}
Hyv\"{a}rinen, A., Shimizu, S., and Hoyer, P.~O. (2008).
\newblock Causal modelling combining instantaneous and lagged effects: an
  identifiable model based on non-{G}aussianity.
\newblock {\em Proceedings of the 25th international conference on machine
  learning (ICML 2008)}, pages 424--431.

\bibitem[Iwasaki and Simon, 1994]{iwasaki:94}
Iwasaki, Y. and Simon, H.~A. (1994).
\newblock Artificial intelligence.
\newblock {\em Causality and model abstraction}, 67:143--194.

\bibitem[Karlebach and Shamir, 2008]{Karlebach2008}
Karlebach, G. and Shamir, R. (2008).
\newblock {Modelling and analysis of gene regulatory networks}.
\newblock {\em Nature Reviews Molecular Cell Biology}, 9(10):770--780.

\bibitem[Koller and Friedman, 2009]{koller:09}
Koller, D. and Friedman, N. (2009).
\newblock {\em Probabilistic Graphical Models: Principles and Techniques}.
\newblock MIT Press.

\bibitem[Korb et~al., 2004]{korb:04}
Korb, K.~B., Hope, L.~R., Nicholson, A.~E., and Axnick, K. (2004).
\newblock Varieties of causal intervention.
\newblock {\em Pacific Rim International Conference on Artificial Intelligence
  (PRICAI 2004)}, pages 322--331.

\bibitem[Kschischang et~al., 2001]{kschis:01}
Kschischang, F., Frey, B., Brendan, J., and Loeliger, H.-A. (2001).
\newblock Factor graphs and the sum-product algorithm.
\newblock {\em IEEE Transactions on Information Theory}, 47:498--–519.

\bibitem[Lattimore et~al., 2016]{lattimore:2016}
Lattimore, F., Lattimore, T., and Reid, M.~D. (2016).
\newblock Causal bandits: Learning good interventions via causal inference.
\newblock In {\em Advances in Neural Information Processing Systems}, pages
  1181--1189.

\bibitem[Lauritzen, 1996]{lauritzen:96}
Lauritzen, S. (1996).
\newblock {\em Graphical Models}.
\newblock Oxford University Press.

\bibitem[Lauritzen and Richardson, 2002]{lauritzen:02}
Lauritzen, S.~L. and Richardson, T.~S. (2002).
\newblock Chain graph models and their causal interpretation.
\newblock {\em Journal of the Royal Statistical Society Series B}, 64:321--361.

\bibitem[LeCun et~al., 2006]{lecun:06}
LeCun, Y., Chopra, S., Hadsell, R., Ranzato, M., and Huang, F.~J. (2006).
\newblock A tutorial on energy-based learning.
\newblock In BakIr, G., Hofmann, T., Sch\"{o}lkopf, B., Smola, A.~J., Taskar,
  B., and Vishwanathan, S., editors, {\em Predicting Structured Data}. MIT
  Press.

\bibitem[Lee and Bareinboim, 2018]{sanghack:2018}
Lee, S. and Bareinboim, E. (2018).
\newblock Structural causal bandits: Where to intervene?
\newblock In {\em Advances in Neural Information Processing Systems}, pages
  2568--2578.

\bibitem[Lee et~al., 2020]{Lee_generalized}
Lee, S., Correa, J., and Bareinboim, E. (2020).
\newblock General transportability – synthesizing observations and
  experiments from heterogeneous domains.
\newblock {\em Proceedings of the AAAI Conference on Artificial Intelligence},
  34(06):10210--10217.

\bibitem[Lei et~al., 2018]{Lei2018}
Lei, J., G'Sell, M., Rinaldo, A., Tibshirani, R.~J., and Wasserman, L. (2018).
\newblock {Distribution-Free Predictive Inference for Regression}.
\newblock {\em Journal of the American Statistical Association},
  113(523):1094--1111.

\bibitem[Leist et~al., 2022]{leist2022mapping}
Leist, A.~K., Klee, M., Kim, J.~H., Rehkopf, D.~H., Bordas, S.~P.,
  Muniz-Terrera, G., and Wade, S. (2022).
\newblock Mapping of machine learning approaches for description, prediction,
  and causal inference in the social and health sciences.
\newblock {\em Science Advances}, 8(42):eabk1942.

\bibitem[Lu et~al., 2023]{lu2023rethinking}
Lu, N., Zhang, T., Fang, T., Teshima, T., and Sugiyama, M. (2023).
\newblock Rethinking importance weighting for transfer learning.
\newblock In {\em Federated and Transfer Learning}, pages 185--231. Springer.

\bibitem[Magliacane et~al., 2018]{magliacane_da}
Magliacane, S., van Ommen, T., Claassen, T., Bongers, S., Versteeg, P., and
  Mooij, J.~M. (2018).
\newblock Domain adaptation by using causal inference to predict invariant
  conditional distributions.
\newblock In {\em Advances in Neural Information Processing Systems},
  volume~31.

\bibitem[Malinsky, 2018]{malinksy:18}
Malinsky, D. (2018).
\newblock Intervening on structure.
\newblock {\em Synthese}, 195:2295–--2312.

\bibitem[McKay, 2003]{mckay:03}
McKay, D. J.~C. (2003).
\newblock {\em Information Theory, Inference, and Learning Algorithms}.
\newblock Cambridge University Press.

\bibitem[Mogensen et~al., 2018]{mogensen:18}
Mogensen, S.~W., Malinsky, D., and Hansen, N.~R. (2018).
\newblock Causal learning for partially observed stochastic dynamical systems.
\newblock {\em Proceedings of the 34th conference on Uncertainty in Artificial
  Intelligence (UAI 2018)}, pages 350--360.

\bibitem[Oberst et~al., 2020]{oberst:20}
Oberst, M., Johansson, F., Wei, D., Gao, T., Brat, G., Sontaga, D., and
  Varshney, K. (2020).
\newblock Characterization of overlap in observational studies.
\newblock {\em $23^{rd}$ International Conference on Artificial Intelligence
  and Statistics (AISTATS 2020)}, pages 788--798.

\bibitem[Ogburn et~al., 2020]{ogburn:20}
Ogburn, E.~L., Shpitser, I., and Lee, Y. (2020).
\newblock Causal inference, social networks and chain graphs.
\newblock {\em Journal of the Royal Statistical Society Series A: Statistics in
  Society}, 183:1659–--1676.

\bibitem[Pearl, 2009]{pearl:09}
Pearl, J. (2009).
\newblock {\em Causality: {M}odels, {R}easoning and {I}nference, 2nd edition}.
\newblock Cambridge University Press.

\bibitem[Pearl and Bareinboim, 2011]{Pearl_Bareinboim_og}
Pearl, J. and Bareinboim, E. (2011).
\newblock Transportability of causal and statistical relations: A formal
  approach.
\newblock {\em Proceedings of the AAAI Conference on Artificial Intelligence},
  25(1):247--254.

\bibitem[Pearl and McKenzie, 2018]{pearl:18}
Pearl, J. and McKenzie, D. (2018).
\newblock {\em The Book of Why}.
\newblock Allen Lane.

\bibitem[Peters et~al., 2016]{peters2016}
Peters, J., Bühlmann, P., and Meinshausen, N. (2016).
\newblock Causal inference by using invariant prediction: identification and
  confidence intervals.
\newblock {\em J. Royal Stat. Soc. Ser. B Methodol.}, 78(5):947--1012.

\bibitem[Pfister et~al., 2019]{pfister2019}
Pfister, N., Bühlmann, P., and Peters, J. (2019).
\newblock Invariant causal prediction for sequential data.
\newblock {\em Journal of the American Statistical Association},
  114(527):1264--1276.

\bibitem[Richardson, 2003]{richardson:03}
Richardson, T. (2003).
\newblock Markov properties for acyclic directed mixed graphs.
\newblock {\em Scandinavian Journal of Statistics}, 30:145--157.

\bibitem[Sachs et~al., 2005]{sachs2005causal}
Sachs, K., Perez, O., Pe'er, D., Lauffenburger, D.~A., and Nolan, G.~P. (2005).
\newblock Causal protein-signaling networks derived from multiparameter
  single-cell data.
\newblock {\em Science}, 308(5721):523--529.

\bibitem[Saengkyongam and Silva, 2020]{saengkyongam:20}
Saengkyongam, S. and Silva, R. (2020).
\newblock Learning joint nonlinear effects from single-variable interventions
  in the presence of hidden confounders.
\newblock {\em $36^{th}$ Conference on Uncertainty in Artificial Intelligence
  (UAI 2020)}.

\bibitem[Sejdinovic et~al., 2013]{dino:13}
Sejdinovic, D., Gretton, A., and Bergsma, W. (2013).
\newblock A kernel test for three-variable interactions.
\newblock {\em Neural Information Processing Systems (NeurIPS)},
  26:1124–--1132.

\bibitem[Shi et~al., 2022]{shi2022assumptions}
Shi, C., Sridhar, D., Misra, V., and Blei, D. (2022).
\newblock On the assumptions of synthetic control methods.
\newblock In {\em International Conference on Artificial Intelligence and
  Statistics}, pages 7163--7175. PMLR.

\bibitem[Shpitser and TchetgenTchetgen, 2016]{shptiser:16a}
Shpitser, I. and TchetgenTchetgen, E. (2016).
\newblock Causal inference with a graphical hierarchy of interventions.
\newblock {\em Annals of Statistics}, 44:2433--2466.

\bibitem[Song and Ermon, 2019]{song:19}
Song, Y. and Ermon, S. (2019).
\newblock Generative modeling by estimating gradients of the data distribution.
\newblock {\em Neural Information Processing Systems (NeurIPS)}, 32.

\bibitem[Spirtes et~al., 2000]{sgs:00}
Spirtes, P., Glymour, C., and Scheines, R. (2000).
\newblock {\em Causation, {P}rediction and {S}earch}.
\newblock Cambridge University Press.

\bibitem[Squires et~al., 2022]{squires2022causal}
Squires, C., Shen, D., Agarwal, A., Shah, D., and Uhler, C. (2022).
\newblock Causal imputation via synthetic interventions.
\newblock In {\em Conference on Causal Learning and Reasoning}, pages 688--711.
  PMLR.

\bibitem[Studen\'{y}, 2051]{studeny:05}
Studen\'{y}, M. (2051).
\newblock {\em Probabilistic conditional independence structures}.
\newblock Springer.

\bibitem[Sussex et~al., 2023]{sussex:23}
Sussex, S., Makarova, A., and Krause, A. (2023).
\newblock Model-based causal {B}ayesian optimization.
\newblock {\em $11^{th}$ International Conference on Learning Representations}.

\bibitem[Tibshirani et~al., 2019]{tibshirani2019}
Tibshirani, R.~J., Foygel~Barber, R., Candes, E., and Ramdas, A. (2019).
\newblock Conformal prediction under covariate shift.
\newblock In {\em Advances in Neural Information Processing Systems},
  volume~32.

\bibitem[Tigas et~al., 2022]{tigasinterventions}
Tigas, P., Annadani, Y., Jesson, A., Sch{\"o}lkopf, B., Gal, Y., and Bauer, S.
  (2022).
\newblock Interventions, where and how? {E}xperimental design for causal models
  at scale.
\newblock In {\em Neural Information Processing Systems (NeurIPS 2022)}.

\bibitem[Vovk et~al., 2005]{vovk2005}
Vovk, V., Gammerman, A., and Shafer, G. (2005).
\newblock {\em Algorithmic Learning in a Random World}.
\newblock Springer, New York.

\bibitem[Winn, 2012]{winn:12}
Winn, J. (2012).
\newblock Causality with gates.
\newblock {\em Proceedings of the Fifteenth International Conference on
  Artificial Intelligence and Statistics, PMLR}, 22:1314--1322.

\bibitem[Zar, 2014]{zar2014spearman}
Zar, J.~H. (2014).
\newblock Spearman rank correlation: overview.
\newblock {\em Wiley StatsRef: Statistics Reference Online}.

\bibitem[Zhang et~al., 2020]{zhang2020}
Zhang, A., Lyle, C., Sodhani, S., Filos, A., Kwiatkowska, M., Pineau, J., Gal,
  Y., and Precup, D. (2020).
\newblock Invariant causal prediction for block {MDP}s.
\newblock In {\em Proceedings of the 37th International Conference on Machine
  Learning}, volume 119, pages 11214--11224. PMLR.

\end{thebibliography}
}
\newpage

\appendix
\section{A Guide to Structure Elicitation and Learning}
\label{appendix:structure}
This section starts with a discussion particularly aimed at methodologists and practitioners who may not feel entirely at ease using a factor graph approach to model causality. 
We then conclude with a brief discussion on how to build IFMs by knowledge elicitation and structure learning. 
For further discussion on the use of alternative independence models in causality, we recommend Section 11 of \cite{dawid:10}. 
Section 12 of the same paper provides further discussion of the meaning of linking intervention variables to random variables in statistical modeling and its DAG interpretations.

\subsection{Background: Extended Conditional Independence}
Our starting point is Dawid's framework for causal inference \citep{dawid:21}. 
Although dubbed ``decision-theoretical'', 
no utility optimization is actually necessary to justify its approach for encoding causal assumptions and deriving their consequences. 
As such, we will drop the ``decision-theoretical'' label and refer to it as the \emph{extended conditional independence} (ECI) approach, formalized in \citep{constantinou:17}. 
The main point is that causal assumptions, either given in the form of independencies in a mutilated DAG or in  distributions of potential outcomes, can directly be framed as ``mere'' statements of conditional independence among random variables and intervention variables. 

As the intervention variables $\sigma_i$ are not random variables, 
this calls for a clarification of what it means to claim $$X_i \indep \sigma_1~|~X_j, \sigma_2$$ 
For what follows, it suffices to interpret this statement as $X_i$ not changing in distribution across different values of $\sigma_1$ when $\sigma_2$ and $X_j$ are fixed/observed at any particular value.

The most basic statement of structure is expressed by a graph \mbox{$X \rightarrow Y$}. 
This graph is intended to communicate that, if we intervene on $X$, the distribution of $Y$ may change; but if we intervene on $Y$, the distribution of $X$ does not change. 
This may feel unsatisfactory, as the graph \mbox{$X \rightarrow Y$} by itself does not communicate any independence constraint.\footnote{The graph may suggest a factorization, and the causal ordering implied by the factorization does matter for  models such as the additive error model \citep{peters2016}. 
However, since these graphs are primarily models of independence, something is still amiss here.}
%

The ECI approach is explicit: we take as primitive the notion of ``intervention on $X$'' and ``intervention on $Y$'' operationalized by a pair of intervention variables $\sigma_x$ and $\sigma_y$ such that
\begin{equation}
\begin{array}{l}
    Y \indep \sigma_x~|~ X\\
    X \indep \sigma_y\\
\end{array}
\label{eq:directed_edge}
\end{equation}
The first independence establishes that once I know \emph{which} value $X$ took, it does not matter \emph{how} $X$ came to be (i.e., the value of $\sigma_x$) \citep{dawid:10}. This is more commonly described as ``lack of unmeasured confounding between $X$ and $Y$'' or \emph{ignorability}. 

The second independence establish that \emph{how} $Y$ comes to be does not change the distribution of $X$. 
This is more commonly described as ``$Y$ does not cause $X$''.
Note that the model does not explicit state that ``$X$ causes $Y$'', just that it's \emph{allowed} to. This mirrors the typical interpretation of a graphical model, by which a graph does not imply \emph{dependencies}. Instead, a graph is defined by its \emph{independencies} \citep{lauritzen:96}. 

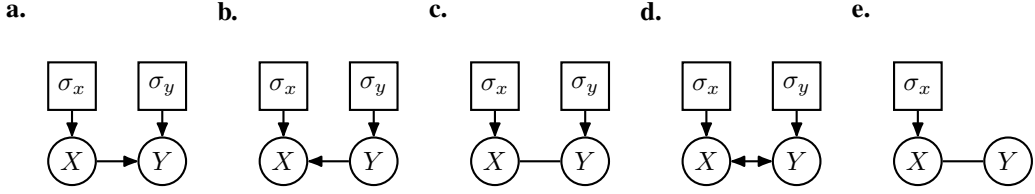
\begin{figure}[!t]
  \raisebox{0.89cm}{
    \hspace*{0.1cm}
    \centering
      \begin{tikzpicture}
        \node[right] (L) at (-1, 2) {\textbf{a.}};
        \node[obs] (X) at (0, 0) {$X$};
        \node[obs] (Y) at (1.2, 0) {$Y$};
        \node[intv] (SX) at (0, 1) {$\sigma_x$};
        \node[intv] (SY) at (1.2, 1) {$\sigma_y$};
        \edge[black]{X}{Y};
        \edge[black]{SX}{X};
        \edge[black]{SY}{Y};
      \end{tikzpicture}%
     
      \hspace*{0.2cm}
      
      \begin{tikzpicture}
        \node[right] (L) at (-1, 2) {\textbf{b.}};
        \node[obs] (X) at (0, 0) {$X$};
        \node[obs] (Y) at (1.2, 0) {$Y$};
        \node[intv] (SX) at (0, 1) {$\sigma_x$};
        \node[intv] (SY) at (1.2, 1) {$\sigma_y$};
        \edge[black]{Y}{X};
        \edge[black]{SX}{X};
        \edge[black]{SY}{Y};
      \end{tikzpicture}%

      \hspace*{0.2cm}
      
      \begin{tikzpicture}
        \node[right] (L) at (-1, 2) {\textbf{c.}};
        \node[obs] (X) at (0, 0) {$X$};
        \node[obs] (Y) at (1.2, 0) {$Y$};
        \node[intv] (SX) at (0, 1) {$\sigma_x$};
        \node[intv] (SY) at (1.2, 1) {$\sigma_y$};
        \uedge[black]{X}{Y};
        \edge[black]{SX}{X};
        \edge[black]{SY}{Y};
      \end{tikzpicture}%
      
      \hspace*{0.2cm}
      
      \begin{tikzpicture}
        \node[right] (L) at (-1, 2) {\textbf{d.}};
        \node[obs] (X) at (0, 0) {$X$};
        \node[obs] (Y) at (1.2, 0) {$Y$};
        \node[intv] (SX) at (0, 1) {$\sigma_x$};
        \node[intv] (SY) at (1.2, 1) {$\sigma_y$};
        \bedge[black]{X}{Y};
        \edge[black]{SX}{X};
        \edge[black]{SY}{Y};
      \end{tikzpicture}%

      \hspace*{0.2cm}
      
      \begin{tikzpicture}
        \node[right] (L) at (-1, 2) {\textbf{e.}};
        \node[obs] (X) at (0, 0) {$X$};
        \node[obs] (Y) at (1.2, 0) {$Y$};
        \node[intv] (SX) at (0, 1) {$\sigma_x$};
        \uedge[black]{X}{Y};
        \edge[black]{SX}{X};
      \end{tikzpicture}%

}
 \vspace{-10mm}
  \caption{The meaning of \mbox{$X \rightarrow Y$}. 
  \textbf{(a)} Expressing it when interventions on both $X$ and $Y$ are defined via variables $\sigma_x$ and $\sigma_y$, and we wish to express that \mbox{$Y \indep \sigma_x~|~ X$} and \mbox{$X \indep \sigma_y$}. 
  \textbf{(b)-(d)} None of these graphs respect the two independence constraints on interest. 
  \textbf{(e)} If $\sigma_y$ is not defined, a graph like this one suffices to encode the remaining constraint \mbox{$Y \indep \sigma_x~|~ X$}.}
 \label{fig:meaning_of_arrow}
 \vspace{-2mm}
\end{figure}

Now, why do we feel compelled to write the edge \mbox{$X \rightarrow Y$}, as in Fig.~\ref{fig:meaning_of_arrow}\textbf{(a)}, as well as the directions from $(\sigma_x, \sigma_y)$ to $(X, Y)$? 
This is because, among all ``canonical graphical models''\footnote{This means the usual machinery of directed, undirected (Markov), mixed and chain graph models \citep{lauritzen:96, richardson:03}, a relatively small but highly interpretable corner in the universe of possible independence models \citep{studeny:05}.}, \emph{that's the only option that we have}. 
To understand that, consider the three variations in Fig.~\ref{fig:meaning_of_arrow}\textbf{(b)-(d)}: each one of them violates one or both of the relations encoded in Eq.~\eqref{eq:directed_edge}. 
That is the sense in which the DAG is justified here: syntactic sugar for Eq.~\eqref{eq:directed_edge}. 
This is particularly emphasized by Fig.~\ref{fig:meaning_of_arrow}\textbf{(e)}: if we do not define $\sigma_y$, the model is defined by the sole constraint \mbox{$Y \indep \sigma_x~|~ X$}. 
In that case, a chain graph with undirected component \mbox{$X - Y$} suffices (and so does the purely undirected structured). 
Any arrows here are for cosmetic purposes.

The same idea applies to \emph{conditional ignorability}: 
if we have some set of covariates $Z$ such that \mbox{$Y \indep \sigma_x~|~X, Z$}, this allows us, for instance, to learn causal effects from observational data. 
That is, if $\sigma_x \in \{\text{``do nothing''}, \text{``$do(X = 0)$''}, \text{``$do(X = 1)$''}\}$, we can obtain  \mbox{$p(y~|~z; \sigma_x = do(X = x))$} from observational data as
\[
\begin{array}{rcll}
p(y~|~z; \sigma_x = do(X = x)) &=& p(y~|~x, z; \sigma_x = do(X = x)) & \text{(since $do(X = x) \Rightarrow X = x$)}\\
&=&p(y~|~x, z; \sigma_x = \text{do nothing}) & \text{(since $Y \indep \sigma_x~|~X, Z$)}\\
&=&p(y~|~x, z) & \text{(conventional notation)}\\
\end{array}
\]

The first line exploits a feature of atomic interventions (\mbox{$do(X = x) \Rightarrow X = x$}) which is not generally available for other types of intervention. 
That is why \mbox{non-atomic} interventions cannot directly be handled by Pearl's \mbox{\textit{do}-calculus} \citep{correa_sigma_calc}. 

As a further example, average treatment effects can be obtained from the backdoor formula/g-formula \citep{pearl:09,hernan:20}, which further requires ``$Z$ not to be caused by $X$''. 
In the ECI framework\footnote{To be clear, the idea of using explicit intervention variables to describe the backdoor adjustment dates back at least to the original graphical formulation of  \cite{sgs:00}. The proof in \cite{pearl:09} also relies on explicit regime variables. ECI formalizes explicitly the role of conditional independence statements that involve non-random variables.}, this is just the further requirement that \mbox{$\sigma_x \indep Z$} (under the interpretation that $\sigma_x$ has ``no causes'' because it comes from a hypothetical agent ``outside the system that generates the data'', this marginal independence can only be explained by $\sigma_x$ ``not causing'' $Z$):
\[
\begin{array}{rcll}
  p(y;\sigma_x = do(X = x)) &=& \sum_z p(y~|~z; \sigma_x = do(X = x))p(z; \sigma_x = do(X = x)) & \\
  &&\text{(standard marginalization)}\\[2pt]
  &=& \sum_z p(y~|~x, z; \sigma_x = do(X = x))p(z)\\&&
  \text{(since $Z \indep \sigma_x$ and $do(X = x) \Rightarrow X = x$)}\\[2pt]
    &=&\sum_z p(y~|~x, z; \sigma_x = \text{do nothing})p(z) &\\&& \text{(since $Y \indep \sigma_x~|~X, Z$)}\\[2pt]
    &=&\sum_z p(y~|~x, z)p(z) &\\&& \text{(conventional notation)}\\
\end{array}
\]

As a matter of fact, the classical axiom of \emph{consistency} that underpins causal reasoning from potential outcomes \citep{hernan:20} (basically, that the potential outcome of $Y$ under intervention \mbox{$do(X = x)$} should match the observed outcome $Y$ if \mbox{$X = x$} i.e. \mbox{$Y_x = Y$} if \mbox{$X = x$}) can be interpreted as the lack of ``fat-hand'' interventions \citep{eberhardt:07}, i.e., that there is some conditioning set $C$ such that \mbox{$\sigma_x \indep Y~|~C$} for any random variable $Y$ other than $X$.

\subsection{Graphical Models and the IFM}

The above discussion indicates that conditional independencies among intervention variables and random variables are the building blocks of a (\mbox{counterfactual-free}, or ``Rung 2''\citep{pearl:18}) causal modeling language. 
However, we need more than that for any practical way of encoding assumptions, as many reasonable models will involve an extremely large number of conditional independencies. 
For instance, even a simple directed Markov chain \mbox{$X_1 \rightarrow X_2 \rightarrow \dots \rightarrow X_p$} involves a super-exponential number of conditional independencies (e.g., $X_p$ is conditionally independence of $X_1$ given any non-empty subset of \mbox{$\{X_2, \dots, X_{p - 1}\}$}). 

\emph{Graphical models} are extremely useful families of independence models that allow for the use of a relatively small number of \emph{local} Markov conditions to describe \emph{global} Markov conditions. 
Moreover, when compared to relying on algebra alone, symbolic algorithms based on graph-theoretical concepts provide ways to derive such implications in a easier and more transparent manner.  
This explains the popularity of graphical models in causal modeling, regardless of the cosmetic appeal of drawing edges. 

In what follows, we will start by contrasting undirected (``Markov'') networks to DAGs, as factor graph models encode the very same families of independencies as undirected graphs (with the extra facility of representing \mbox{low-order} interactions on top of independence constraints).
For instance, in a DAG, the local Markov condition is a variable being conditional independent of its \mbox{non-descendants} given its parents; the global Markov condition is anything entailed by \mbox{d-separation} \citep{pearl:09,sgs:00}. 
See \cite{lauritzen:96} for many classical results.

Creating an independence (graphical) model requires \mbox{trade-offs}, as not every family of conditional independencies can be easily cast in \mbox{graph-theoretical} terms \citep{studeny:05}. 
One common example is the (chordless) simple cycle of length $4$: 
the independencies encoded in this undirected graph \mbox{$X_1 - X_2 - X_3 - X_4 - X_1$} cannot be represented by any DAG, even one with the same adjacencies, such as, \mbox{$X_1 \rightarrow X_2 \rightarrow X_3 \leftarrow X_4 \leftarrow X_1$}.
(The converse is also true, the independencies encoded by this  ``diamond structure'' DAG have no correspondence to any undirected graph.) 
It is out of our scope to get into any of the fine details of such differences, see \cite{lauritzen:96} for details. 
Instead, we will focus on some broad aspects relevant to causal inference.

DAGs (with or without hidden variables) are by far the most common type of graphical model for expressing causality. 
There are different ways of explaining this appeal, of which we highlight  \emph{``marginal independencies''} and \emph{``explaining away''}.

\textbf{Marginal Independencies.} A connected undirected graph implies no marginal independendencies. 
Yet, marginal independencies lie behind the claim that ``the future does not cause the past''. 
This is illustrated in Fig.~\ref{fig:meaning_of_arrow}\textbf{(a)} by interpreting $Y$ as encoding events that happen after the events encoded by $X$, and hence an assumption of \mbox{$X \indep \sigma_y$} is desirable. 

\textbf{Explaining Away.} 
This well-known phenomenon is illustrated by independencies that get destroyed by conditioning upon further evidence. 
For example, if \mbox{$X_3 = f(X_1, X_2)$}, then even if \mbox{$X_1 \indep X_2$}, it is clear that knowing also the value of $X_3$ will change the support of $X_1$ given $X_2$. 
More generally, \mbox{$p(x_1, x_2, x_3) = p(x_1)p(x_2)p(x_3~|~x_1, x_2)$} means that \mbox{$p(x_1, x_2) = g(x_1)h(x_2)$} for some $g(\cdot), h(\cdot)$, but \mbox{$p(x_1, x_2~|~x_3) \propto g(x_1)h(x_2)k_{x_3}(x_1, x_2)$} for some $k_{x_3}(\cdot, \cdot)$ which does not factorize in general. 
This also applies to combinations of random variables and intervention variables. If ``$X$ does not cause $Z$'' and we operationalize that by \mbox{$\sigma_x \indep Z$} in the DAG $\sigma_x \rightarrow X \leftarrow Z$, it is possible to have this independence destroyed by conditioning on $X$, since $p(z~|~x; \sigma_x)$ will in general be different if $\sigma_x$ is ``do nothing'' (where $Z$ is allowed to vary with different values of $X$) compared against \mbox{$\sigma_x = do(X = x)$} (where $Z$ is independent of $X$). 
That is, in this example we have \mbox{$Z \not\!\perp\!\!\!\perp \sigma_x~|~X$}, in the sense that $p(z~|~x; \sigma_x)$ is a \mbox{non-trivial} function of $\sigma_x$ even if \mbox{$\sigma_x \indep Z$}.
%

We claim that neither of the two properties above are particularly well-motivated in the snapshot sampling process illustrated in Figure~\ref{fig:process}. 
Given the snapshot, some explaining away could indeed happen between an intervention variable $\sigma$ and \mbox{pre-treatment}/``Past'' variables (Fig.~\ref{fig:process}\textbf{(b)}) that happen to be recorded.
However, this can be simply accounted for by including these \mbox{pre-treatment} variables as conditioning variables within the IFM factors. 

For longitudinal studies, where marginal independencies do matter (``the future doesn't cause the past'') \textit{and} we do happen to make multiple snapshots (as opposed to ``contemporaneous DAG structures'', as in \citep{aapo:08}), it is not a technical challenge to extent the IFM to a chain graph structure \citep{lauritzen:02} with factor graph undirected components. 
We leave this development for future work.

To summarize, if marginal independencies and explaining away are not of particular relevance to the problem at hand, then we recommend the IFM as a family of independence models, particularly in light of its very simple local Markov condition (i.e., in the corresponding undirected graph implied by the factor structure, ``a variable is independent of its non-neighbors given its neighbors''). 
This comes to life especially in models motivated by equilibrium equations. 
For instance, 
consider the following example from Section 3.1.1 of \cite{blom:23}: 
\[
\begin{array}{rl}
f_I: & X_I - U_I = 0\\
f_D: & U_1(X_I - X_O) = 0\\
f_P: & U_2(gU_3X_D - X_P) = 0\\
f_O: & U_4(U_5I_KX_P - X_O) = 0
\end{array}
\]
where $f$ denotes differential equations at equilibrium, 
$X$ denotes observed random variables, 
$U$ denotes (mutually independent) latent variables,
and $I$ denotes an intervention indicator. 

After marginalizing the $U$ variables, we ``get'' an IFM (although, this is not the whole story, as other \mbox{non-graphical} constraints may take place given the particular equations. \cite{blom:23} discusses $U_I$ being marginally independent of $I_K$ on top of the above). 
In general, energy-based models are to be interpreted as conjunctions of soft-constraints, the factor graph is one implication of a system of stochastic differential equations (SDE), and interventions denote change to particular constraints. 

A SDE model can indeed have \emph{other} assumptions on top of the factorization, and parameters which carry particular meaningful interpretations.
Nonetheless, this fits well with our claims that the IFM is a ``minimalist'' family of models in terms of structural assumptions --- a reasonably conservative direction to follow; particularly, when the dynamics of many natural phenomena cannot be (currently) measured at individual level, as it's the case of much of cell biology data, and hence writing down a full SDE model may be more inspirational than scientifically grounded. See \cite{lauritzen:02} for further elaborations on some of these ideas.

It is worth highlighting that, as long as the timing of the measurements is \mbox{well-defined} and consistently respected by the \mbox{real-world} sampling procedure, there is no need to wait for a system to get into equilibrium in order for an IFM to be applicable. 
\cite{dash:05} discusses how ``causal structure'' changes depending on which equations have reached equilibrium at any particular point in time --- after all, the process as a function of time is non-stationary until (and if) it reaches equilibrium. 
Importantly, intervention generalization here should \emph{not} be interpreted as extrapolating to different, unsampled, time points in a \mbox{non-stationary} process.

Finally, we note that the ECI framework as described by \cite{constantinou:17} is relatively restricted in the definition of statements such as $$\sigma_i \indep \sigma_j ~|~X_r, \sigma_s,$$ that is, claims of independence between sets of intervention variables. 
The focus there is on \emph{variation independence}, which roughly speaking can be interpreted as the range of possible values for $\sigma_i$ not depending on $\sigma_j$.\footnote{As it would not be the case, for instance, if a choice of $\sigma_j$ corresponds to removing a condition or resource required for carrying out an action encoded by some values of $\sigma_i$. This happens, for example, in resource allocation problems, where $\sigma_i$ and $\sigma_j$ correspond to resource allocation decisions limited by some budget constraint.} 

In contrast, we interpret statements $$\sigma_i \indep \sigma_j ~|~X, \sigma_{\backslash ij},$$ where $\sigma_{\backslash ij}$ are all intervention indicators other than $\sigma_i$ and $\sigma_j$, by merely linking it to the factorization $$p(x; \sigma_i, \sigma_j, \sigma_{\backslash ij}) \propto g(x, \sigma_i, \sigma_{\backslash ij})h(x, \sigma_j, \sigma_{\backslash ij}),$$ for some functions $g(\cdot)$ and $h(\cdot)$. Going from pairwise independence to setwise independence is defined here by the usual graphical criterion of pairwise independence for the product space of the sets implying setwise independence.

The above does not mean a genuine factorization of \mbox{$p(x; \sigma_i, \sigma_j, \sigma_{\backslash ij})$} as a function of $\sigma$, as the normalizing constant will in general depend on all regime indicators (but not on the data). 
It however denotes the difference between Fig.~\ref{fig:graph_differences}\textbf{(b)} and Fig.~\ref{fig:graph_differences}\textbf{(c)}: the latter is a chain graph with an undirected component that does not suggest factorization over $\sigma_1$, $\sigma_2$ and $\sigma_3$ for any given $x$, while the factor graph explicitly encodes that.

\subsection{Structure Elicitation and Learning}

Having agreed that an intervention factor model is the natural choice under the scenarios described above, 
the remaining consideration
is: 
how to extract structural knowledge from an expert or algorithm. 

Simply put, removing edges from $\sigma$ to $X$ (or among $X$) \emph{should be business as usual} once we understand that structure follows from conditional independencies among random variables and regime indicators, with the global Markov condition being that of a factor graph model. 
An expert who is ready to answer conditional ignorability questions of the type \mbox{$Y \indep \sigma_x~|~X, Z$}, and/or plain \mbox{consistency-like} questions to judge whether \mbox{$\sigma_x \indep Y~|~C$} for some $C$ and \mbox{$Y \neq X$}, should have no qualms about answering general questions for interventions that don't have a particular \mbox{single-variable} target. 
In particular, the notion of ``removing edges'' from $\sigma$ to $X$ (that is, forbidding any factor to include particular combinations of intervention and random variables) should follow from knowledge about the domain. 
Indeed,  impossibility of some direct connections is presented in all sorts of systems, from physical ones (including spatial systems \citep{akbari:23}) to social ones (e.g., \citep{ogburn:20}).

Independencies of the type \mbox{$\sigma_i \indep \sigma_j~|~X, \sigma_{\backslash ij}$} are better understood as the lack of particular interactions. 
\mbox{Non-linear} multivariate models such as \mbox{log-linear} models have for long been understood as defining hierarchies of interactions within the allowed probabilistic dependencies \citep{bishop:74}. 
Likewise, analysis of variance (ANOVA) is predicated on the idea that \mbox{lower-order} interactions can suffice to model a variety of empirical phenomena. 
Judging whether a set of random variables (``soft constraint'') should be regulated by interactions of particular intervention variables, conditional on all other variables, is knowledge akin to judging interactions in ANOVA or \mbox{log-linear} models, and it has a long tradition in multivariate analysis, dating back at least to the work of Ising on statistical mechanics \citep{mckay:03}.

However, none of that is to say that the work of structure elicitation is straightforward; it is definitely not. 
With this in mind, we now conclude this section with broad ideas on structure learning for IFMs.
%
Indeed, structure learning can aid the process of structure elicitation. 
An early method for structure learning of probabilistic factor graph models is described by \cite{abbeel:06}, and it is our intent to provide a fully detailed account of structure learning for IFMs in future work.

There is a close link between classical DAG learning algorithms and algorithms for undirected models, using variants of the faithfulness assumption \citep{sgs:00}. 
In particular, akin to the initial stage of the PC algorithm \citep{sgs:00}, we can start with a fully connected undirected graph and remove edges, creating a factor graph model out of the cliques remaining after a step that removes edges. 

We can remove edges between random variables and edges between random variables and intervention variables, by querying an independence oracle under a particular regime $\sigma^0$, which we assume all other regimes should be faithful to. 
For instance, this takes place when $\sigma^0$ is an ``observational regime'' as defined by an unperturbed system, and any independence among random variables is assumed be carried over to other regimes. 
Likewise, varying one entry in $\sigma$ and assessing (conditional) equality in distribution for particular random variables will remove undirected edges between $\sigma$ and $X$ by assuming that they are also be unnecessary under any other configuration of the unchanged variables. 

Finally, we note that edges ``within'' $\sigma$ variables, and interactions in general, are less straightforward to deal with. 
For any clique remaining in the current undirected graph, one possibility is to test whether the distribution of this clique given all other variables provides the same goodness-of-fit (by statistical significance) with or without particular interactions. 
Statistical power may be an issue, see \cite{dino:13} for a discussion on nonparametric testing of three-way interactions. 
An alternative is to adopt a blanket assumption to remove higher-order interactions if there is no evidence against lower-order interactions across models fit separately in each of the $\mathcal D^i$ datasets collected.
%

\newpage
\section{Proofs of Identifiability and Further Examples}
\label{appendix:proof}

This section presents results concerning Theorems \ref{theorem:main} and \ref{theorem:algebraic}. 
We start with some background with textbook definitions, followed by proofs and examples for the decomposable graph, and concluding with proofs for the purely algebraic case. 
In particular, 
the decomposable case sheds light on how to hierarchically structure products and ratios of $\mathbb P(\Sigma_{\train})$. 
Among other uses, this theoretically suggests which regimes could be added to $\Sigma_{\train}$ in order to reduce the estimation error coming from particular product/ratios that are required to identify larger marginal distributions.

\paragraph{Background.} 
A \emph{decomposition} of an undirected graph $\mathcal G$ is formed from a partition of its vertices into a triplet $(A, B, C)$ where $C$ is a complete subgraph (i.e., a clique) and separates $A$ from $B$. 
The decomposition is called \emph{proper} if both $A$ and $B$ are non-empty. 
Moreover, an undirected graph $\mathcal G$ is called \emph{decomposable} if it is complete or, failing that, it has a proper decomposition into a triplet $(A, B, C)$, where the subgraph of $\mathcal G$ with vertices $A \cup B$ and the subgraph with vertices $B \cup C$ are both decomposable.

A \emph{junction tree} $\mathcal T$ of a decomposable graph $\mathcal G$ is a tree where each vertex $V_i$ is labeled with the elements of a unique maximal clique from $\mathcal G$ (hence, this type of vertex is sometimes called a hypervertex), so that \mbox{$V_i \cap V_j$} denotes the corresponding intersection among sets of vertices in $\mathcal G$. 
Edges \mbox{$V_i - V_j$} of $\mathcal T$ are graphically represented with labels denoting the intersection \mbox{$V_i \cap V_j$} (hence, these edges are also sometimes called hyperedges). 
A junction tree must have a \emph{running intersection} property: any intersection \mbox{$V_i \cap V_j$} must be contained in all vertices in the (unique) path between $V_i$ and $V_j$ in $\mathcal T$.


If $\mathcal G_{\sigma(\mathcal I)}$ is decomposable, there exists at least one junction tree compatible with it. Let $\mathcal T$ be one of them. Without loss of generality, pick an arbitrary vertex of $\mathcal T$ to be the root and direct edges away from it to create a directed tree out of the junction tree, so that we can assume $\mathcal T$ to be directed. We will prove identifiability by an induction argument that starts at the leaves of the directed junction tree, moving towards the (unique) parent of any particular vertex child in the induction step. 

\paragraph{Definitions and notation.} In what follows, we use $V_k$ to denote  a vertex in $\mathcal T$. By abuse of notation, depending on context, $V_k$ is also used to denote the corresponding intervention variables $\sigma_{F_k}$ in the original factor graph.

Let $\sigma^{[Z(w)]}$ be a particular instantiation of $\sigma$, where \mbox{$Z \subseteq [d]$} and \mbox{$\sigma_{Z} = w$} (possibly a vector), with the remaining entries of $\sigma$ being zero. 
For instance, if \mbox{$d = 3$}, \mbox{$Z = \{2, 3\}$}, and \mbox{$w = (2, 1)$}, then \mbox{$\sigma^{[Z(w)]} = (0, 2, 1)$}. 
To avoid subsequently heavy notation, from this point on we will use $\sigma^{[Z(\star)]}$ to denote $\sigma^{[Z(\sigma^\star_Z)]}$, that is, the vector of assignments that we obtain by setting to zero all entries of $\sigma^\star$ which are not in $Z$.

We use $ch(k)$ to denote the set of children of vertex \mbox{$V_k$ in $\mathcal{T}$}, and $V_{\pi(k)}$ to denote its (unique) parent, if $V_k$ is not the root vertex. 
Also, let $D_k$ denote the union of the intervention variables contained in at least one descendant of $V_k$ in $\mathcal{T}$, remembering that by convention $V_k$ is also a descendant of itself. 
Finally, let \mbox{$B_k := D_k \cap V_{\pi(k)}$}, the set of intervention variables common to both $D_k$ and $V_{\pi(k)}$. 
This means that, by the running intersection property of junction trees, $B_k$ separates \mbox{$A_k := D_k \backslash B_k$} from the rest of $\sigma$ in the \mbox{$\sigma$-graph} $\mathcal{G}_{\sigma(\mathcal{I})}$.

\paragraph{Proof of Theorem \ref{theorem:main}.} 
To simplify the proof, assume without loss of generality that no entry in $\sigma^\star$ is zero. 
To see this, if \mbox{$\sigma_i^\star = 0$}, consider the factors $k$ containing $\sigma_i$ as being constants in $\sigma_i$, with scope $S_k$ being redefined as \mbox{$S_{k'} := S_k \backslash \{\sigma_i\}$}. 
$\Sigma_{\train}$ in this redefined space still satisfies the assumption of having entries spanning all possible values for $\sigma_{S_{k'}}$ while holding the remaining intervention variables at the background level of $0$. 
Likewise, as identifiability will be shown pointwise for a given $\sigma^{\star}$ (where the categorical labels for the intervention values are arbitrary symbols), we can assume all entries $\sigma_i^\star$ as being equal, and equal to $1$.

We define a \emph{message} from vertex $V_k$ to its parent $V_{\pi(k)}$ as
\begin{equation}
m_k^x := \frac{p(x; \sigma^{[D_k(\star)]})}{p(x; \sigma^{[B_k(\star)]})},
\label{eq:message}
\end{equation}
\noindent and state that
\begin{equation}
p(x; \sigma^{[D_k(\star)]}) \propto
p(x; \sigma^{[F_k(\star)]})\prod_{V_{k'} \in ch(k)} m_{k'}^x,
\label{eq:update}
\end{equation}
with the product over $ch(k)$ defined to be $1$ if $ch(k) = \emptyset$. 

We will show how Eq.~\eqref{eq:message} can be recursively identified from a message scheduling that starts from the leaves and propagates messages towards the root of $\mathcal T$. 
We will also show how Eq.~\eqref{eq:update} holds.

Let $V_k$ be a leaf of $\mathcal T$. 
Then \mbox{$D_k = F_k$} and $p(x; \sigma^{[F_k(\star)]})$ is identified as it is part of $\Sigma_{\train}$, showing that Eq.~\eqref{eq:update} holds for the leaf vertices of $\mathcal T$. 
Likewise, message $m_k^x$ is identifiable for leaf vertices, as both $D_k$ and $B_k$ are contained in $F_k$.

Let $V_k$  now be an internal vertex of $\mathcal T$, and assume that Equations \eqref{eq:message} and \eqref{eq:update} are identified for all of its proper descendants. 
As all entries of $\sigma^\star$ are assumed to be $1$, it will be useful to define $g_k := f_k(x; \sigma_{F_k} = 1)$. 
Also define \mbox{$h_k := f_k(x; \sigma_{F_k \cap B_k} = 1, \sigma_{F_k \backslash B_k} = 0)$} and \mbox{$z_k := f_k(x; \sigma_{F_k} = 0)$}.

Since $\mathcal T$ is a junction tree, $D_k$ can be partitioned into sets $D_{k'}$, where \mbox{$V_{k'} \in ch(k)$}, or otherwise there would be a violation of the running intersection property. 
Let $Q(k')$ be the set of all indices of factors $q$ where \mbox{$F_q \subseteq D_{k'}$}. 
Let
\[
\mathcal Q_{k'} := \prod_{q \in Q(k')} \frac{g_q}{h_q}.
\]
We can multiply and divide $\mathcal Q_{k'}$ by the product of all factors $z_q$, where \mbox{$F_q \cap D_{k'} = \emptyset$}. 
This implies
\[
\mathcal Q_{k'} \propto \frac{p(x; \sigma^{[D_{k'}(\star)]})}{p(x; \sigma^{[B_{k'}(\star)]})} = m_{k'}^x.
\]
Moreover, the product
\[
\frac{\prod_{q: F_q \cap D_{k} = \emptyset} z_q}{\prod_{q: F_q \cap D_{k} = \emptyset} z_q} \times \frac{f_k(x; \sigma_{F_k} = 1)}{f_k(x; \sigma_{F_k} = 1)} \times 
\frac{f_{\pi(k)}(x; \sigma_{B_k} = 1, \sigma_{F_{\pi(k)} \backslash B_k} = 0)}{f_{\pi(k)}(x; \sigma_{B_k} = 1, \sigma_{F_{\pi(k)} \backslash B_k} = 0)} \times
\prod_{k': V_{k'} \in ch(k)} \mathcal Q_{k'}
\]
is such that the numerator is proportional to $p(x; \sigma^{[D_k(\star)]})$ and the denominator is proportional to $p(x; \sigma^{[F_k(\star)]})$. 
To see this, note that the numerator sets the $\sigma_{F_j}$ variables for all factors $F_j$ in a coherent way such that entries in $D_k$ are set to 1 while everything else is set to zero (entries in $D_k$ may still appear in $V_{\pi(k)}$, as \mbox{$D_k \cap F_{\pi(k)} = B_k$}, is possibly non-empty. 
Hence, we set to $1$ those entries in $F_{\pi(k)}$ which are in $B_k$, explaining the appearance of the $f_{\pi(k)}$ factors in the expression above).

This implies
\[
\frac{p(x; \sigma^{[D_k(\star)]})}{p(x; \sigma^{[F_k(\star)]})} \propto 
\prod_{V_{k'} \in ch(k)} m_{k'}^x,
\]
from which Eq.~\eqref{eq:update} follows from quantities previously identified, and as such it identifies $p(x; \sigma^{[D_k(\star)]})$. 

To build message $m_k^x$, all that remains is $p(x; \sigma^{[B_k(\star)]})$. 
However, \mbox{$B_k \subseteq F_k$}, and since $\Sigma_{\train}$ contains the distribution $p(x; \sigma^{[H_k(\star)]})$ for all \mbox{$H_k \subseteq F_k$}, this is also identified. The required identifiability of $p(x; \sigma^\star)$ follows from propagating these messages all the way up to the root of $\mathcal T$. $\Box$

\begin{figure}[h]
\hspace{0.8in}
  \raisebox{0.2cm}{
  \begin{tikzpicture}
    \node[intv] (S1) at (0, 1) {$\sigma_1$};
    \node[intv] (S2) at (1.2, 1) {$\sigma_2$};
    \node[intv] (S3) at (2.4, 1) {$\sigma_3$};
    \node[intv] (S4) at (3.6, 1) {$\sigma_4$};
    \node[intv] (S5) at (4.8, 1) {$\sigma_5$};
    \node[intv] (S6) at (6.0, 1) {$\sigma_6$};
    \node[intv] (S7) at (7.2, 1) {$\sigma_7$};
    \node[intv] (S8) at (8.4, 1) {$\sigma_8$};
    \node[fact] (F1) at (0.6, 0) {\textcolor{white}{$f_1$}};
    \uedge{S1}{F1};
    \uedge{S2}{F1};
    \node[fact] (F2) at (1.8, 2) {\textcolor{white}{$f_2$}};
    \uedge{S2}{F2};
    \uedge{S3}{F2};
    \node[fact] (F3) at (3.0, 0) {\textcolor{white}{$f_3$}};
    \uedge{S2}{F3};
    \uedge{S5}{F3};
    \node[fact] (F4) at (3.0, 2) {\textcolor{white}{$f_4$}};
    \uedge{S3}{F4};
    \uedge{S4}{F4};
    \node[fact] (F5) at (4.8, 2) {\textcolor{white}{$f_5$}};
    \uedge{S4}{F5};
    \uedge{S6}{F5};
    \node[fact] (F6) at (6.6, 2) {\textcolor{white}{$f_6$}};
    \uedge{S6}{F6};
    \uedge{S7}{F6};
    \node[fact] (F7) at (6.0, 0) {\textcolor{white}{$f_7$}};
    \uedge{S4}{F7};
    \uedge{S8}{F7};
  \end{tikzpicture}%
  }
 \caption{The factor graph used as an illustration of the technique in the proof of Theorem \ref{theorem:main}.}
 \label{fig:example_proof_1}
\end{figure}
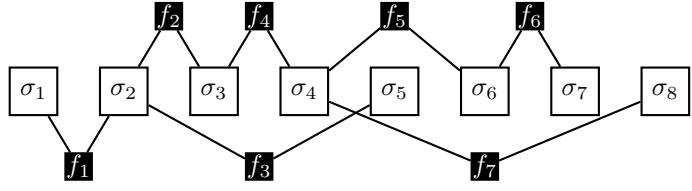

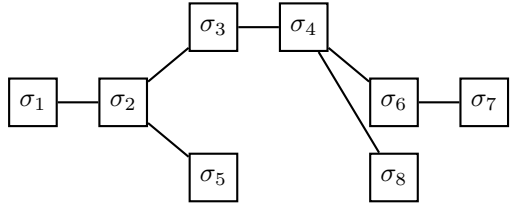
\begin{figure}[h]
 \hspace{1.2in}
  \raisebox{0.2cm}{
  \begin{tikzpicture}
    \node[intv] (S1) at (0, 1) {$\sigma_1$};
    \node[intv] (S2) at (1.2, 1) {$\sigma_2$};
    \node[intv] (S3) at (2.4, 2) {$\sigma_3$};
    \node[intv] (S5) at (2.4, 0) {$\sigma_5$};
    \node[intv] (S4) at (3.6, 2) {$\sigma_4$};
    \node[intv] (S6) at (4.8, 1) {$\sigma_6$};
    \node[intv] (S8) at (4.8, 0) {$\sigma_8$};
    \node[intv] (S7) at (6.0, 1) {$\sigma_7$};
    \uedge{S1}{S2};
    \uedge{S2}{S3};
    \uedge{S2}{S5};
    \uedge{S3}{S4};
    \uedge{S4}{S6};
    \uedge{S4}{S8};
    \uedge{S6}{S7};
  \end{tikzpicture}%
  }
 \caption{The $\sigma$-graph corresponding to the factor graph in Figure~\ref{fig:example_proof_1}.}
 \label{fig:example_proof_2}
\end{figure}

\begin{figure}[h]
 \hspace{1.2in}
  \raisebox{0.2cm}{
  \begin{tikzpicture}
    \node[hobs] (S12) at (0, 0) {$(1, 2)$};
    \node[hobs] (S25) at (2.4, 0) {$(2, 5)$};
    \node[hobs] (S67) at (4.8, 0) {$(6, 7)$};
    \node[hobs] (S67) at (4.8, 0) {$(6, 7)$};
    \node[hobs] (S48) at (7.2, 0) {$(4, 8)$};
    \node[hintv] (I12) at (0, 1.2) {$2$};
    \node[hintv] (I25) at (2.4, 1.2) {$2$};
    \node[hintv] (I67) at (4.8, 1.2) {$6$};
    \node[hintv] (I48) at (7.2, 1.2) {$4$};
    \node[hobs] (S23) at (1.2, 2.4) {$(2, 3)$};
    \node[hobs] (S46) at (6.0, 2.4) {$(4, 6)$};
    \node[hintv] (I23) at (2.4, 3.6) {$3$};
    \node[hintv] (I46) at (4.8, 3.6) {$4$};
    \node[hobs] (S34) at (3.6, 4.8) {$(3, 4)$};
    \uedge{S34}{I23};
    \uedge{S34}{I46};
    \edge{I23}{S23};
    \edge{I46}{S46};
    \edge{I23}{S23};
    \uedge{S23}{I12};
    \uedge{S23}{I25};
    \uedge{S46}{I67};
    \uedge{S46}{I48};
    \edge{I12}{S12};
    \edge{I25}{S25};
    \edge{I67}{S67};
    \edge{I48}{S48};
  \end{tikzpicture}%
  }
 \caption{A (directed) junction tree corresponding to the undirected graph in Figure~\ref{fig:example_proof_2}.}
 \label{fig:example_proof_3}
 \vspace{-2mm}
\end{figure}
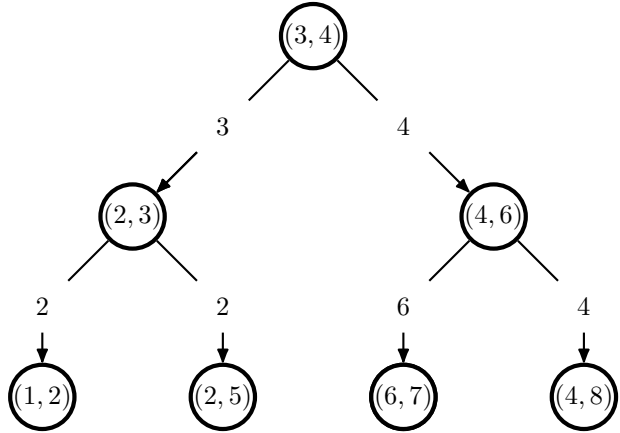

\paragraph{Example.} 
We now solve the example shown in Figures~\ref{fig:example_proof_1}--\ref{fig:example_proof_3}. 
The IFM itself is given by 
\[
p(x; \sigma) \propto f_1(x; \sigma_1, \sigma_2)
f_2(x; \sigma_2, \sigma_3)
f_3(x; \sigma_2, \sigma_5)
f_4(x; \sigma_3, \sigma_4)
f_5(x; \sigma_4, \sigma_6)
f_6(x; \sigma_6, \sigma_7)
f_7(x; \sigma_4, \sigma_8),
\]
\noindent where all intervention variables are binary, 
and we our goal is to generate the test regime where all intervention variables are set to $1$, i.e., \mbox{$\sigma_1 = \sigma_2 = \dots = \sigma_8 = 1$}. 
%
For reference, this means considering the following sets implied by the factorization above:
\[
\begin{array}{ccc}
F_1 = \{1, 2\} & D_1 = \{1, 2\} & B_1 = \{2\}\\
F_2 = \{2, 3\} & D_2 = \{1, 2, 3, 5\} & B_2 = \{3\}\\
F_3 = \{2, 5\} & D_3 = \{2, 5\} & B_3 = \{2\}\\
F_4 = \{3, 4\} & D_4 = \{1, 2, 3, 4, 5, 6, 7, 8\} & B_4 = \emptyset\\
F_5 = \{4, 6\} & D_5 = \{4, 6, 7, 8\} & B_5 = \{4\}\\
F_6 = \{6, 7\} & D_6 = \{6, 7\} & B_6 = \{6\}\\
F_7 = \{4, 8\} & D_7 = \{4, 8\} & B_7 = \{4\}\\
\end{array}
\]

To illustrate how message passing will work in this example, 
let's introduce some symbols so that the steps are easier to follow. 
Let $g_{ij}^{11}$ represent a factor with \mbox{$\sigma_i = \sigma_j = 1$}. 
For instance, \mbox{$g_{12}^{11} = f_1(x; \sigma_1 = 1, \sigma_2 = 1)$}. 
This is slightly redundant compared to the notion in the proof (which uses ``$g_1$'' to denote \mbox{$f_1(x; \sigma_1 = 1, \sigma_2 = 1)$}), but the redundancy of the superscripts will hopefully make it easier to visualize the logic in the steps that follow.

Likewise, let $h_{ij}^{01}$ and $h_{ij}^{10}$ denote assignments \mbox{$(\sigma_i, \sigma_j) = (0, 1)$} and \mbox{$(\sigma_i, \sigma_j) = (1, 0)$}, respectively. 
Finally, let $z_{ij}^{00}$ denote the respective factor with assignment \mbox{$\sigma_i = \sigma_j = 0$}.

We will use expressions such as $p(x; [ij])$ to denote \mbox{$p(x; \sigma_i = 1, \sigma_j = 1, \sigma_{1:8 \backslash \{i, j\}} = 0)$} to make the notation simpler.

The messages at the leaves are
\[
\begin{array}{rcll}
m_1^x &=& p(x; [12]) / p(x; [2]) & (D_1 = \{1, 2\}, B_1 = \{2\})\\
m_3^x &=& p(x; [25]) / p(x; [2]) & (D_3 = \{2, 5\}, B_3 = \{2\})\\
m_6^x &=& p(x; [67]) / p(x; [6]) & (D_6 = \{6, 7\}, B_6 = \{6\})\\
m_7^x &=& p(x; [48]) / p(x; [4]) & (D_7  = \{4, 8\}, B_7 = \{4\})\\
\end{array}
\]
It can be readily verified that all of these are identifiable from $\Sigma_{\train}$, as all non-zero assignments are contained within some factor.

Now, let's pass messages to $(2, 3)$ using formula (\ref{eq:update}). 
To see how it is applicable, start from
\[
\begin{array}{rcl}
\displaystyle
m_1^x \times m_3^x &=& 
\displaystyle
\frac{p(x; [12])}{p(x; [2])}\frac{p(x; [25])}{p(x; [2])}\\
\\
&\propto&
\displaystyle
\frac{g_{12}^{11}h_{23}^{10}h_{25}^{10}z_{34}^{00}z_{46}^{00}z_{67}^{00}z_{48}^{00}}
{h_{12}^{01}h_{23}^{10}h_{25}^{10}z_{34}^{00}z_{46}^{00}z_{67}^{00}z_{48}^{00}} \times
\frac{h_{12}^{01}h_{23}^{10}g_{25}^{11}z_{34}^{00}z_{46}^{00}z_{67}^{00}z_{48}^{00}}
{h_{12}^{01}h_{23}^{10}h_{25}^{10}z_{34}^{00}z_{46}^{00}z_{67}^{00}z_{48}^{00}}
\end{array}
\]
Now, we multiply and divide it by the factor of $(2, 3)$ and its parent $(3, 4)$ evaluated at $(\sigma_2, \sigma_3, \sigma_4) = (1, 1, 0)$, and reorganize the numerator and denominator:
\[
\begin{array}{rcl}
\displaystyle
m_1^x \times m_3^x &=& 
\displaystyle
\frac{p(x; [12])}{p(x; [2])}\frac{p(x; [25])}{p(x; [2])}\\
\\
&\propto&
\displaystyle
\frac{g_{12}^{11}\bcancel{h_{23}^{10}}\bcancel{h_{25}^{10}}\bcancel{z_{34}^{00}}\bcancel{z_{46}^{00}}\bcancel{z_{67}^{00}}\bcancel{z_{48}^{00}}}
{h_{12}^{01}\bcancel{h_{23}^{10}}\bcancel{h_{25}^{10}}\bcancel{z_{34}^{00}}\bcancel{z_{46}^{00}}\bcancel{z_{67}^{00}}\bcancel{z_{48}^{00}}} \times
\frac{\bcancel{h_{12}^{01}}\bcancel{h_{23}^{10}}g_{25}^{11}\bcancel{z_{34}^{00}}z_{46}^{00}z_{67}^{00}z_{48}^{00}}
{\bcancel{h_{12}^{01}}\bcancel{h_{23}^{10}}h_{25}^{10}\bcancel{z_{34}^{00}}z_{46}^{00}z_{67}^{00}z_{48}^{00}} \times \frac{g_{23}^{11}}{g_{23}^{11}}
\times \frac{h_{34}^{10}}{h_{34}^{10}}\\
\\
&=&
\displaystyle
\frac{g_{12}^{11}g_{23}^{11}g_{25}^{11}h_{34}^{10}z_{46}^{00}z_{67}^{00}z_{48}^{00}}
{h_{12}^{01}g_{23}^{11}h_{25}^{10}h_{34}^{10}z_{46}^{00}z_{67}^{00}z_{48}^{00}}\\
\\
&\propto&
\displaystyle
\frac{p(x; [1235]}{p(x; [23])}.
\end{array}
\]
As $p(x; [23])$, $m_1^x$ and $m_3^x$ have been previously identified, from the above we get the update for $p(x; [1235])$ per Eq.~\eqref{eq:update}, pointing out that indeed \mbox{$D_2 = \{1, 2, 3, 5\}$} and \mbox{$F_2 = \{2, 3\}$}.

To construct the message $m_2^x$ the factor $(2, 3)$ needs to pass to its own parent $(3, 4)$, we also need the corresponding $p(x; \sigma^{[B_2(\star)]})$, which in the example notation is $p(x; [3])$. 
But as \mbox{$B_2 = \{3\}$} is contained in \mbox{$F_2 = \{2, 3\}$}, and this will be the case for all $(B_k, F_k)$ pairs, by assumption $\Sigma_{\train}$ will contain $p(x; [3])$. Therefore, we identified $m_2^x$.

The steps for $(4, 6)$ and $(3, 4)$ follow identical, if somewhat tedious, reasoning. 

\newpage

\vspace{30pt}
\paragraph{Proof of Theorem \ref{theorem:algebraic}.} 
Sufficiency follows immediately from the fact that, under Eq~\eqref{eq:algebraic} being satisfied, the \mbox{PR-transformation} \mbox{$\prod_{i = 1}^tp(x; \sigma^i)^{q_i}$} is equivalent to
\begin{equation}
\displaystyle
\prod_{k = 1}^l \prod_{\sigma_{F_k}^v \in \mathbb D_k} f_k(x_{S_k}; \sigma_{F_k}^v)^{\sum_{i = 1: \sigma_{F_k}^i = \sigma_{F_k}^v}^t q_i} = 
\prod_{k = 1}^l f_k(x_{S_k}; \sigma_{F_k}^\star) \propto p(x; \sigma^\star),
\label{eq:algebraic_sufficiency}
\end{equation}
\noindent for all $x$.

For almost-everywhere necessity, let \mbox{$z_{kv} := \log f_k(x_{S_k}; \sigma_{F_k}^v)$}. Taking the logarithm on both sides of the equality in Eq.~\eqref{eq:algebraic_sufficiency}, we have
\[
\displaystyle
\sum_{k = 1}^l \sum_{\sigma_{F_k}^v \in \mathbb D_k} z_{kv}\left(\sum_{i = 1: \sigma_{F_k}^i = \sigma_{F_k}^v}^t q_i\right) = 
\sum_{k = 1}^l z_{k^\star},
\]
\noindent which implies
\[
\displaystyle
\sum_{k = 1}^l z_{k^\star}\left(\sum_{i = 1: \sigma_{F_k}^i = \sigma_{F_k}^\star}^t q_i\right) + 
\sum_{k = 1}^l \sum_{\sigma_{F_k}^v \in \mathbb D_k \backslash \{\star\}} z_{kv}\left(\sum_{i = 1: \sigma_{F_k}^i = \sigma_{F_k}^v}^t q_i\right) = 0.
\]

As no $z_{k^\star}$ appears in the second term of the expression above, the only way for this equality to hold without \mbox{$\{q_1, \dots, q_t\}$} satisfying Eq.~\eqref{eq:algebraic} is if constrains tie together the different $z_{k^\star}$. 
For any reasonable continuous measure by which the parameters of such functions are free to be chosen from (say, as draws of a multivariate Gaussian), this will be a set of measure zero. $\Box$

\paragraph{Discussion.} 
As a corollary it is implied that, similar to the conditions of Theorem \ref{theorem:main}, we need to have at least one train condition $\sigma^i$ for every possible combination of $\sigma_{F_k}$, for each $F_k$. 
To see why, imagine if the example of Figure~\ref{fig:algebraic_method} we did not have condition 4, that is, \mbox{$(\sigma_1, \sigma_2, \sigma_3) = (1, 1, 0)$} is left out. 
This means that there is nothing to be added in the column $f_1^{11}$, and the sum \smash{\mbox{$\sum_{i = 1: \sigma_{F_1}^i = (1, 1)}^t q_i$}} evaluates to $0$, implying \mbox{$0 = 1$}.

\section{More on Elicitation, Testability and Experimental Design}
\label{appendix:testability}

As mentioned before, the main result shows that the factorization over $X$ is unimportant for identifiability, which may be surprising. 
However, it is important to remember that identifiability and testability are two different concepts. 
While Figure~\ref{fig:graph_differences}\textbf{(b)} has testable implications of conditional independence, testing factorizations may require more intervention levels than the minimal set implied by Theorem~\ref{theorem:main}. 
In particular, if we have a model \smash{\mbox{$p(x; \sigma) \propto \prod_{k = 1}^l f_k(x_k; \sigma_k)$}}, we may be able to identify the model by singleton experiments spanning the range of each $\sigma_i$ individually, however, this does not mean we can falsify this factorization with just this data. 

In general, our advice for graph construction is akin to any causal modeling exercise: apply independence constraint tests and interaction tests where applicable (see \citep{dino:13} for an example of nonparametric \mbox{three-way} interaction test), but untestable conditions (under the available data) can be used if there is a sensible theoretical justification for it. 
This means expert assessment of the lack of direct dependency between an intervention variable and particular random variables, and of the split of $\sigma$ into sets $F_k$ from postulated lack of interactions among intervention variables when causing particular random variables. 
Although not necessarily always the case, we anticipate that in general this exercise will imply a factorization over the random variables too.

Also of interest is understanding which minimal size $\Sigma_{\train}$ should have in order to identify a particular test regime. 
This is straightforward to answer in the decomposable case: simply ensure that the regimes used in the messages of the message passing scheme are available in the training set. 
A simple iterative algorithm can list the required messages for a target regime $\sigma^\star$. 
For example, with binary treatments, a \mbox{$\sigma$-graph} without any edges (no interaction of intervention variables in a same factor), and the goal of identifying \emph{all} combinations of interventions, this is simply \mbox{$d + 1$}, where $d$ is the number of intervention variables (this follows from having the baseline regime plus one regime where a single intervention variable is set to $1$). 

For non-decomposable graphs, we can triangulate the corresponding \mbox{$\sigma$-graph} and run the same procedure defined for decomposable graphs to provide an upper bound on number of training conditions and a superset of conditions. 
We can run a greedy procedure to iteratively remove redundant entries in $\Sigma_{\train}$ by proposing candidate training regimes to be removed and testing whether the PR condition for the regime $\sigma^\star$ of interest is still satisfied.

What if the cardinality $\aleph_i$ of some $\sigma_i$ is very high? Without smoothness assumptions, getting a reasonable dose-response pattern with few evaluations of $\sigma_i$ is clearly impossible regardless of any method --- this is true even for a single intervention variable in the $[0, 1]$ interval where (say) $p(x; \sigma)$ jumps arbitrarily as we sweep the values of $\sigma$ in $[0, 1]$. 
\emph{With} smoothness assumptions, we can simply elicit a grid of values for the intervention variables, ask conditions for the identifiability of those, and fill up the remaining potential functions/expected outcome values of interest via whatever smoothing procedure we deem appropriate (from potential functions which are smooth functions of $\sigma$ or via partial identification procedures, see e.g. \cite{higbee:23}). There is no free lunch.

\section{Proof of Predictive Coverage Result}

\paragraph{Proof of Theorem \ref{thm:conform}.} \label{appx:conformal}
\citet{vovk2005} introduced a distribution-free procedure for computing prediction intervals with guaranteed finite sample coverage, under the assumption that training and test data are exchangeable. 
\citet{Lei2018} proposed a more computationally tractable version that they called the ``split conformal'' method, and derived a novel upper bound on conformal coverage. 
We review some fundamental results.

Consider the regression setting with \mbox{$X \in \mathcal{X} \subseteq \mathbb{R}^d$} and \mbox{$Y \in \mathcal{Y} \subseteq \mathbb{R}$}. 
We partition the data into equal-sized subsets $\mathcal{I}_1, \mathcal{I}_2$, using the former for model training and the latter for computing conformity scores. 
For instance, we may fit a model $\widehat{f}(x)$ to estimate $\mathbb{E}[Y \mid x]$ using samples from $\mathcal{I}_1$ and consider the score function \mbox{$s^{(i)} = |y^{(i)} - \widehat{f}(x^{(i)})|$} for $i \in \mathcal{I}_2$. 
Let $\widehat{\tau}$ be the $q^{\text{th}}$ smallest value in $S$, with \mbox{$q = \lceil (n/2 + 1) (1 - \alpha) \rceil$}. Define \mbox{$\widehat{C}(x) = \widehat{f}(x) \pm \widehat{\tau}$}. (We assume symmetric errors for convenience; the result can easily be modified by invoking the appropriate quantiles of the residual distribution.)

\begin{theorem}[Split conformal inference \citep{Lei2018}.] 
Fix a target level \mbox{$\alpha \in (0,1)$}.
If \mbox{$(x^{(i)}, y^{(i)}), i \in [n]$}, are exchangeable, then for any new $n+1$ from the same distribution:
\begin{align*}
    \mathbb{P}\big(Y^{(n+1)} \in \widehat{C}(X^{(n+1)})\big) \geq 1 - \alpha.
\end{align*}
Moreover, if scores have a continuous joint distribution, then the upper bound on this probability is \mbox{$1 - \alpha + 2/(n+2)$}.
\end{theorem}

\citet{tibshirani2019} extend this result beyond exchangeable data by introducing the notion of \textit{weighted exchangeability}. 
We call random variables \mbox{$V_1, \dots, V_n$ }\textit{weighted exchangeable}, with weight functions \mbox{$w_1, \dots, w_n$}, if their joint density can be factorized as:
\begin{align*}
    f(v_1, \dots, v_n) = \prod_{i=1}^n w_i(v_i) \cdot g(v_1, \dots, v_n),
\end{align*}
where $g$ does not depend on the ordering of its inputs, i.e., $g$ is permutation invariant. This entails the following lemma.
\begin{lemma}[Weighted exchangeability \citep{tibshirani2019}.]
  Let \mbox{$Z_i \sim P_i, i \in [n]$}, be independent draws, where each $P_i$ is absolutely continuous with respect to $P_1$, for \mbox{$i \geq 2$}. Then \mbox{$Z_1, \dots, Z_n$} are weighted exchangeable, with weight functions $w_1 = 1$ and \mbox{$w_i = \text{d}P_i / \text{d}P_1, i \geq 2$}. 
\end{lemma}
This allows us to generalize the conformal guarantee to weighted exchangeable distributions. 
Let $\widetilde{w}^{(i)}(x)$ denote a rescaled version of the weight function, such that weights sum to $n$. 
The original paper does not use the split conformal approach, but we adapt the result below. First, we reweight the empirical scores to create the new distribution \mbox{$\sum_i \widetilde{w}^{(i)}(x) \cdot \delta(s)^{(i)}$}, where $\delta$ denotes the Dirac delta function. 
Now, let $\widehat{\tau}(x)$ be the $q^{\text{th}}$ smallest value in \mbox{$\sum_i \widetilde{w}^{(i)}(x) \cdot \delta(s)^{(i)}$}, with $q$ defined as above. 
Then, we construct the weighted conformal band \mbox{$\widehat{C}_w(x) = \widehat{f}(x) \pm \widehat{\tau}(x)$} for all \mbox{$x \not\in \mathcal{I}_1$}. 

\begin{theorem}[Split weighted conformal inference \citep{tibshirani2019}.] 
Fix a target level \mbox{$\alpha \in (0,1)$}.\\
If \mbox{$(x^{(i)}, y^{(i)})$}, are weighted exchangeable with weight functions \mbox{$w^{(i)}, i \in [n]$}, then for any new $n+1$:
\begin{align*}
    \mathbb{P}\big(Y^{(n+1)} \in \widehat{C}_w(X^{(n+1)})\big) \geq 1 - \alpha.
\end{align*}
Moreover, if scores have a continuous joint distribution, then the upper bound on this probability is \mbox{$1 - \alpha + 2/(n+2)$}.
\end{theorem}

Our case is somewhat trickier, as we do not have access to data $X$ from the unobserved environment $\sigma^\star$ and our regime variables are not random, so ratios such as \mbox{$p(\sigma^a) / p(\sigma^b)$} are undefined. 
However, we can use a similar reweighting strategy based on likelihood ratios of the form \mbox{$p(x^{k(i)}; \sigma^\star) / p(x^{k(i)}; \sigma^k)$} to ensure that conformity scores satisfy weighted exchangeability with respect to any target regime. 
This works because we observe the mediators $x$ for each conformity score $s$, and assume identifiability of the relevant likelihood ratios via previous Theorems~\ref{theorem:main} and/or~\ref{theorem:algebraic}. 
Thus our conformal bands are functions of $\sigma$, not $X$, and our result is simply a special case of the split weighted conformal inference theorem. $\Box$

\section{Covariate Shift Method}
\label{appendix:covariate_shift}

As IPW may have large variance, one alternative is to use covariate shift regression \citep{lu2023rethinking}. 
In particular, for each test regime $\sigma^\star$, we provide a customize estimate of \mbox{$f_y(x) := \mathbb E[Y | X]$}.

As before, we combine data from all training regimes \mbox{$\mathcal D^1, \dots, \mathcal D^t$}, but reweighting then according to (the estimated) $p(x; \sigma^\star)$. We propose minimizing the following objective function,
\[
\mathcal L_y(\theta_y) := \sum_{i = 1}^t \sum_{j = 1}^{n_i}
(y^{ij} - f_y(x^{ij}; \theta_y))^2 w^{ij^\star},
\]
where $w^{ij^\star}$ is an estimate of \mbox{$p(x^{ij}; \sigma^\star)/p(x^{ij}; \sigma^i)$}, and all the training data regimes are weighted equally given that \mbox{$\sum_{j = 1}^n w^{ij^\star} = 1$} for all $i$. 
As done with the IPW method, this ratio is taken directly from the likelihood function of the deep energy-based model we described for the direct method.

When generating an estimate $\widehat \mu_{\sigma^\star}$, we just apply the same idea as in the direct method, where samples from the estimated $p(x; \sigma^\star)$ are generated by Gibbs sampling, so that we average $f_y(x; \widehat \theta_y)$ over these samples.

As we are averaging over $f_y(X)$ instead of considering predictions at each realization of $X$, the main motivation for covariate shift here is to improve on IPW by substituting the use of $Y^{ij}$ as the empirical plug-in estimate of $\mathbb E[Y^{ij}~|~x^{ij}]$ with a smoothed version of it given by a shared learned $f_y(X^{ij}; \theta_y)$. 

However, in our experiments, this covariate shift method was far too slow when considering the cost over the entire $\Sigma_{\test}$ (as expected, given that the output model is fitted again for every test regime) and did not show concrete advantages compared to the direct method.

\section{Pseudolikelihood Method}
\label{appendix:learning}

The pseudo-loglikelihood function \mbox{$p\mathcal L(\theta; \mathcal D^1, \dots, \mathcal D^t)$} is given by
\[
p\mathcal L(\theta; \mathcal D^1, \dots, \mathcal D^t) :=
\sum_{i = 1}^t \sum_{j = 1}^{n_i} \sum_{r = 1}^m
\log p_\theta(x^{i(j)}_r~|~x^{i(j)}_{\backslash r}; \sigma^i),
\]
\noindent where $x^{i(j)}_r$ is the $r$-th variable of the $j$-th data point in dataset $\mathcal D^i$, with $n_i$ being the number of data points in $\mathcal D^i$. Vector $x^{i(j)}_{\backslash r}$ for the same data point is composed of all other random variables but the $r$-th variable.

The log-conditional distribution $\log p_\theta(x^{i(j)}_r~|~x^{i(j)}_{\backslash r}; \sigma^i)$ is given by
\[
\displaystyle
\log p_\theta(x^{i(j)}_{k}~|~x^{i(j)}_{\backslash k}; \sigma^{i}) =
\sum_{k = 1}^{l} \phi_{k, \sigma_{F_k}}(x_{S_k}^{i(j)}) - 
\log\left\{\sum_{x'_{r}} 
\exp\left(\sum_{k = 1}^{k} \phi_{k,\sigma_{F_{k}}}({x'}^{ i(j)}_{S_{k}})\right)\right\},
\]
\noindent where the second term on the right-hand side is the log-normalizing constant summing all possible values $x'_r$ of the $r$-th random variable. 
Here, ${x'}^{ i(j)}_{S_{k}}$ is the vector obtained by substituting the value of $x_k$ within data point $j$ of dataset $i$ with $x'_k$, prior to selecting the subvector corresponding to $S_k$.

The above assumes that all variables are discrete. As described in the main text, we discretize our variables in an uniform grid, preserving the magnitude information. The parameterization $\theta$ is the same regardless of the number of discretization levels, so that this number is chosen basically by the computational considerations of performing the sum over $x'_k$ in the log-normalizing constant. Finer discretizations preserve more information but increase this cost.

\newpage
\section{Experimental Details}
\label{appendix:experimental_details}

In this section, we present further experimental details for Section \ref{sec:experiments}, including the full setup for the datasets (Sachs and DREAM), the oracular simulators (Causal-DAG and Causal-IFM) that generate the ground-truth $X$ and $Y$, and the training procedures we used in the experiments. 
 
\subsection{Datasets}
\label{appendix:datasets}

\paragraph{Sachs (et al.) dataset.} 
The original Sachs et al.~study \citep{sachs2005causal} consisted of 14 different datasets collected under different compound perturbations in single-cell systems measured by 11 protein/lipid concentrations. 
Perturbations can be described in terms of binary intervention variables, labeled by the associated compound. 
For instance, condition $pma$ describes the introduction or not of phorbol 12-myristate 13-acetate. 
Among all  perturbations, $pma$ and $b2camp$ are entangled with $cd3cd28$ (this means, for instance, that $pma = 1$ or $b2camp = 1$ imply $cd3cd28off=1$). 
Hence, we ignore these two experimental setups, and all remaining datasets are collected under  $cd3cd28 = 1$ so that it can be considered as an implicit condition not modeled explicitly with a separate intervention variable. 

Other conditions implying unresolved entanglements were not considered, in particular the uses of $icam$-$2$ and $ly294002$. 
The remaining datasets are listed in Table~\ref{appendix:tab:Sachs}. 
Assumptions about each intervention targeting a single protein in the network are taken from \cite{sachs2005causal}. 
In summary, the original Sachs et al.~data used to train the simulator contains samples from 5 (1 ``baseline'' plus 4 ``perturbed'') different regimes, and each data sample has 11 variables. 

Since each intervention is considered as a binary value (0 for no perturbation and 1 for perturbation), this gives us a total of $2^4=16$ combinatorial possibilities, with 5 in $\Sigma_{\train}$. Hence, we need a way of establishing a (synthetic) ground truth for the $16 - 5 = 11$ possible test conditions, which we explain in Section \ref{appendix:simulators}.

\paragraph{DREAM dataset.} 
The DREAM challenges include a series of problems for causal inference in protein networks \cite{greenfield2010dream4}. We generate data based on a known \textit{E.~coli} sub-network with 10 nodes, and consider that each random variable $X_i$ has a corresponding interventional variable $\sigma_i$. We use the GeneNetWeaver simulator\footnote{\url{https://gnw.sourceforge.net}} to generate this data, under "InSilicoSize10-Ecoli1" from the "DREAM3\_In-Silico\_Size\_10" task and there is no further data selection process as in the Sachs case. The simulation is based on a series of predefined ODEs and SDEs. For each data regime, a single data sample is collected with a random seed initialization with an otherwise exact similar simulation setting for that particular regime. Following \citep{tigasinterventions}, we gather the data sample once it reaches its equilibrium state and repeat this process as many times as the sample size is required. In summary, this provides us with a dataset consisting of 11 (1 baseline plus 10 perturbation) regimes. As we are interested in combinations of 10 binary indicator variables $\sigma_i$, not directly provided in the original DREAM simulator, we had to create our own ground-truth synthetic model based on samples from the 11 regimes we can obtain from DREAM.

\begin{table}[h]
\small
\centering
\caption{Details for the Sachs et al. datasets used for our first batch of intervention generalization experiments. Data files can be downloaded from the website of the original reference \citep{sachs2005causal}, with the name described below. Column \emph{Target $X$ node} describes the theoretical direct connection (as given by \citep{sachs2005causal}) between the perturbation and 11-dimensional system described by a vector of 11 random variables $X$, with condition $cd3cd28$ always present and affecting all variables, and hence interpreted as a targeting none. As described in Section \ref{appendix:simulators}, we encode each regime as a 11-dimensional binary vector, and display them in the last column. A Julia notebook exemplifying the pre-processing of this data and a Julia script outlining a complete pipeline of batch simulated experiments comparing methods is provided in the supplementary material. 
}
\begin{tabular}{l|l|l|l}
\hline
File name & Target $X$ Node & Data Regime & Corresponding $\sigma$\\
\hline
cd3cd28.xls & None (background condition) &  Regime 0 & [0, 0, 0, 0, 0, 0, 0, 0, 0, 0, 0] \\
cd3cd28+aktinhib.xls & Variable 7 & Regime 1 & [0, 0, 0, 0, 0, 0, 1, 0, 0, 0, 0] \\
cd3cd28+g0076.xls & Variable 9 & Regime 2 & [0, 0, 0, 0, 0, 0, 0, 0, 1, 0, 0] \\
cd3cd28+psitect.xls & Variable 4 & Regime 3 & [0, 0, 0, 1, 0, 0, 0, 0, 0, 0, 0] \\
cd3cd28+u0126.xls & Variable 2 & Regime 4 & [0, 1, 0, 0, 0, 0, 0, 0, 0, 0, 0] \\
\hline
\end{tabular}
\label{appendix:tab:Sachs}
\end{table}

\subsection{Oracular Simulators}
\label{appendix:simulators}

Both Sachs and DREAM come with a ground truth DAG (either defined by expert domain knowledge or motivated by physical systems dynamics). We used each of the DAGs to construct the associated IFMs. 
To further explain: in a DAG, the joint probability distribution can be factorized based on the local Markov condition \citep{lauritzen:96}, where a single factor is defined by a vertex and its parents; this suggests are least one IFM, with the factorization following from interpreting each child-parents factor as a black-box (i.e., not normalized by the child) positive function of these variables. 
Graphically, this is known as the \emph{moralization} step of ``marrying the parents'' followed by the dropping of directions in order to create an undirected Markov network \citep{lauritzen:96}. 
We use this to define a factor graph model without stating that this would be the best representation for the corresponding data. 
It relaxes the DAG assumption (i.e., it removes some of the independence constraints encoded in the DAG) and could be refined by adding other constraints (such as breaking the factors into products of reduced sets of variables), which we do not attempt. See Figure \ref{fig:graph_sachs} for an example with the Sachs et al.~model. 
Approaches such as \citep{abbeel:06} could be used to refine this structure, if so desired.

Given two theoretical constructions and the respective parameterizations they use, this suggests \emph{two} ways of building ground-truth simulator models to generate ground truth data $X$, which we now explain.

Common to both ground-truth simulators is the fitting of a postulated causal structure to real data (either Sachs or DREAM). 
Prior to fitting, we scale the data of each study so that the respecive ``merged empirical distributions'', defined by taking the union of all respective datasets collected under all available training regimes, have empirical mean of zero and empirical variance of 1 for each measured random variable. 
This does not mean any given variable in any given training regime will have zero empirical mean and unit variance, but pragmatically it helps to control having variables with disparate scales. 
For the Sachs data, we also take the logarithm of each random variable prior to standardization.

\subsubsection{Causal DAG Ground-Truth}
\label{appendix:experiment:dag_generator}

\begin{figure}[h]
  \raisebox{0.89cm}{
  \begin{tabular}{c}
  \begin{tikzpicture}
    \node[right] (L) at (-1, 2) {\textbf{a.}};
    \node[obs] (X1) at (0.0, 2) {$X_1$};
    \node[obs] (X2) at (2.4, 2) {$X_2$};
    \node[obs] (X3) at (4.8, 2) {$X_3$};
    \node[obs] (X4) at (7.2, 2) {$X_4$};
    \node[obs] (X5) at (9.8, 4) {$X_5$};
    \node[obs] (X6) at (9.8, 2) {$X_6$};
    \node[obs] (X7) at (12.6, 2) {$X_7$};
    \node[obs] (X8) at (6.0, 4) {$X_8$};
    \node[obs] (X9) at (0.0, 4) {$X_9$};
    \node[obs] (X10) at (4.8, 6) {$X_{10}$};
    \node[obs] (X11) at (7.2, 6) {$X_{11}$};

    \node[intv] (S2) at (2.4, 0) {$\sigma_2$};
    \node[intv] (S4) at (7.2, 0) {$\sigma_4$};
    \node[intv] (S7) at (12.6, 0) {$\sigma_7$};
    \node[intv] (S9) at (0.0, 6) {$\sigma_9$};

    \edge[black]{X1}{X2};
    \edge[->, bend right=30]{X2}{X6};
    \edge[black]{X3}{X4};
    \edge[black]{X3}{X9};
    \edge[black]{X4}{X9};
    \edge[black]{X5}{X3};
    \edge[black]{X5}{X4};
    \edge[black]{X5}{X7};
    \edge[black]{X6}{X7};
    \edge[black]{X8}{X1};
    \edge[black]{X8}{X2};
    \edge[black]{X8}{X6};
    \edge[black]{X8}{X7};
    \edge[black]{X8}{X10};
    \edge[black]{X8}{X11};
    \edge[black]{X9}{X1};
    \edge[black]{X9}{X2};
    \edge[black]{X9}{X8};
    \edge[black]{X9}{X10};
    \edge[black]{X9}{X11};
    \edge[black]{S2}{X2};
    \edge[black]{S4}{X4};
    \edge[black]{S7}{X7};
    \edge[black]{S9}{X9};
  \end{tikzpicture}%
  \\
  \vspace{0.5cm}
  \\
  \begin{tikzpicture}

    \node[right] (L) at (-1, 2) {\textbf{b.}};

    \node[obs] (X1) at (0.0, 0) {$X_1$};
    \node[obs] (X2) at (1.2, 0) {$X_2$};
    \node[obs] (X3) at (2.4, 0) {$X_3$};
    \node[obs] (X4) at (3.6, 0) {$X_4$};
    \node[obs] (X5) at (4.8, 0) {$X_5$};
    \node[obs] (X6) at (6.0, 0) {$X_6$};
    \node[obs] (X7) at (7.2, 0) {$X_7$};
    \node[obs] (X8) at (8.4, 0) {$X_8$};
    \node[obs] (X9) at (9.6, 0) {$X_9$};
    \node[obs] (X10) at (10.8, 0) {$X_{10}$};
    \node[obs] (X11) at (12.0, 0) {$X_{11}$};

    \node[fact] (F1) at (0.0, 3) {\textcolor{white}{$f_1$}};
    \node[fact] (F2) at (1.2, 3) {\textcolor{white}{$f_2$}};
    \node[fact] (F3) at (2.4, 3) {\textcolor{white}{$f_3$}};
    \node[fact] (F4) at (3.6, 3) {\textcolor{white}{$f_4$}};
    \node[fact] (F5) at (4.8, 3) {\textcolor{white}{$f_5$}};
    \node[fact] (F6) at (6.0, 3) {\textcolor{white}{$f_6$}};
    \node[fact] (F7) at (7.2, 3) {\textcolor{white}{$f_7$}};
    \node[fact] (F8) at (8.4, 3) {\textcolor{white}{$f_8$}};
    \node[fact] (F9) at (9.6, 3) {\textcolor{white}{$f_9$}};
    \node[fact] (F10) at (10.8, 3) {\textcolor{white}{$f_{10}$}};
    \node[fact] (F11) at (12.0, 3) {\textcolor{white}{$f_{11}$}};

    \node[intv] (S2) at (1.2, 4) {$\sigma_2$};
    \node[intv] (S4) at (3.6, 4) {$\sigma_4$};
    \node[intv] (S7) at (7.2, 4) {$\sigma_7$};
    \node[intv] (S9) at (9.6, 4) {$\sigma_9$};

    \edge[black]{S2}{F2};
    \edge[black]{S4}{F4};
    \edge[black]{S7}{F7};
    \edge[black]{S9}{F9};

    \uedge[black]{X1}{F1};
    \uedge[black]{X8}{F1};
    \uedge[black]{X9}{F1};
    \uedge[black]{X2}{F2};
    \uedge[black]{X1}{F2};
    \uedge[black]{X8}{F2};
    \uedge[black]{X9}{F2};
    \uedge[black]{X3}{F3};
    \uedge[black]{X5}{F3};
    \uedge[black]{X4}{F4};
    \uedge[black]{X3}{F4};
    \uedge[black]{X5}{F4};
    \uedge[black]{X5}{F5};
    \uedge[black]{X6}{F6};
    \uedge[black]{X6}{F6};
    \uedge[black]{X2}{F6};
    \uedge[black]{X8}{F6};
    \uedge[black]{X7}{F7};
    \uedge[black]{X5}{F7};
    \uedge[black]{X6}{F7};
    \uedge[black]{X8}{F7};
    \uedge[black]{X8}{F8};
    \uedge[black]{X9}{F8};
    \uedge[black]{X9}{F9};
    \uedge[black]{X3}{F9};
    \uedge[black]{X4}{F9};
    \uedge[black]{X10}{F10};
    \uedge[black]{X8}{F10};
    \uedge[black]{X9}{F10};
    \uedge[black]{X11}{F11};
    \uedge[black]{X8}{F11};
    \uedge[black]{X9}{F11};
    
  \end{tikzpicture}
  
  \end{tabular}
 }
 \caption{Causal structures used for building the synthetic ground-truth models for the Sachs et al. \citep{sachs2005causal} data. The name of the random variables are $CD3CD28off$, $ICAM$-$2$, $Akt$-$inhibitor$, $G0076$, $Psitectorigenin$, $U0126$, $LY294002$, $PMA$, $B2camp$, with more details given in the companion Julia notebook.
 \textbf{(a)} A directed acyclic graph (DAG) for the Sachs et al. process, with intervention vertices representing intervention variables. 
 \textbf{(b)} The interventional factor graph, inspired by the DAG, which we use in our synthetic ground-truth simulator. This is done by creating a factor for each child and parent set from the postulated DAG. These independence models are not equivalent. 
 The point is \emph{not} to provide an exact model, but to build a synthetic ground truth with parameters calibrated by real data instead of arbitrarily sampled, and with independence constraints and factorizations that do not contradict a given expert assessment (as the factor graph contains \emph{fewer} independence assumptions than the DAG, not more).}
 \label{fig:graph_sachs}
\end{figure}
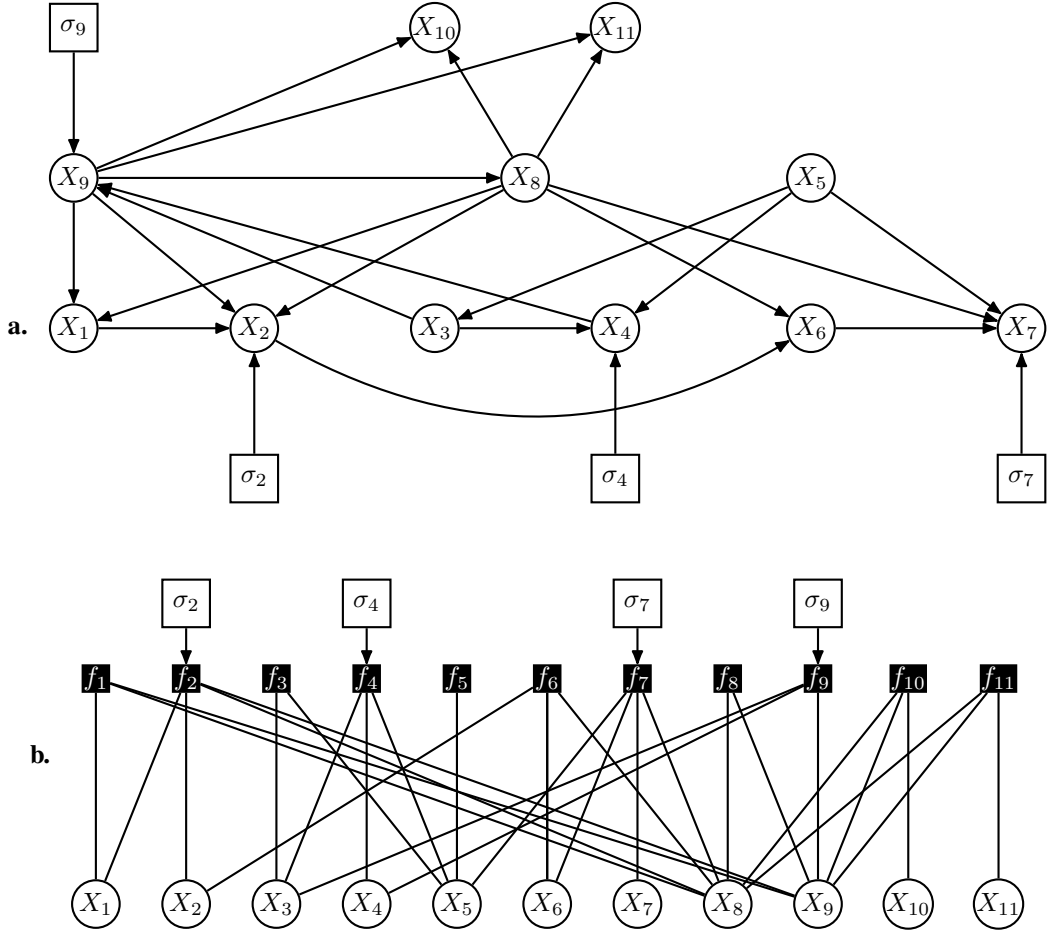

The first ground-truth simulator is implied by the respective causal DAG model. 
The DAG for each study are shown in Figures~\ref{fig:graph_sachs}\textbf{(a)} and \ref{fig:graph_dream}\textbf{(a)}. 
The factorization comes from the structure of the DAG and can be rewritten as follows:
\begin{equation}
p(x; \sigma) = \prod_{k = 1}^l p(x_{k} ~|~ \textbf{pa}(x_{k}); \sigma_{F_k}),
\label{eq:causal_dag}
\end{equation}
where, $l$ is the total number of random variables, $p(x_{k}~|~ \textbf{pa}(x_{k}); \sigma_{F_k})$ is the conditional density function for $x_k$, $\sigma_{F_k}$ is the regime indicator subvector for the intervention variables which are parents of $x_k$ in the DAG, and $\textbf{Pa}(x_{k})$ refers to the random variables which are parents of $x_{k}$. 

For parameterizing the causal DAG model family, we assume a heteroscedastic conditionally Gaussian formulation. This can be represented by the equation $$X_{k} = f_k(\textbf{pa}(X_{k}), \sigma_k) + g_k(\textbf{pa}(X_{k}), \sigma_k) \times \epsilon_k, \epsilon_k \sim \mathcal{N}(0, 1).$$ 
Here, each $f_k$ and $g_k$ are multilayer perceptrons (MLPs) with 10 hidden units and the role of $\sigma_k$ is just a switch: for each value of $\sigma_k$, we pick one independent set of parameters for the MLP mapping $\textbf{pa}(X_{k})$ to the real line. To learn the parameters in functions $f_k$ and $g_k$, maximum likelihood is used. Further details are provided in the companion Julia code.

\newpage

\subsubsection{IFM Ground-Truth} 

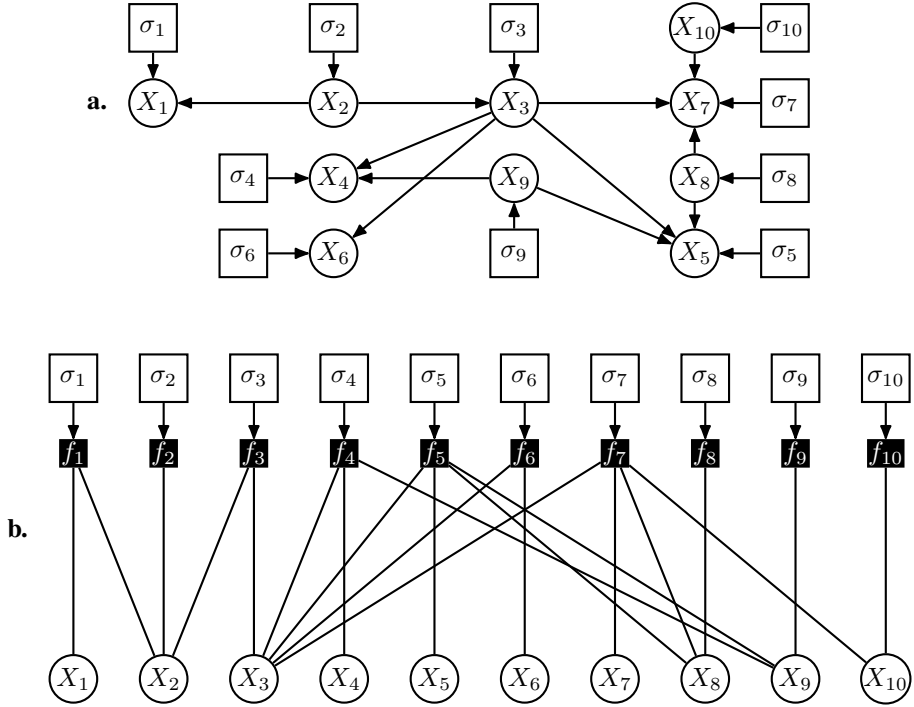
\begin{figure}[h]
  \raisebox{0.89cm}{
  \begin{tabular}{c}
  \hspace{0.15in}
  \begin{tikzpicture}
    \node[right] (L) at (-1, 2) {\textbf{a.}};
    \node[obs] (X1) at (0.0, 2) {$X_1$};
    \node[obs] (X2) at (2.4, 2) {$X_2$};
    \node[obs] (X3) at (4.8, 2) {$X_3$};
    \node[obs] (X4) at (2.4, 1) {$X_4$};
    \node[obs] (X5) at (7.2, 0) {$X_5$};
    \node[obs] (X6) at (2.4, 0) {$X_6$};
    \node[obs] (X7) at (7.2, 2) {$X_7$};
    \node[obs] (X8) at (7.2, 1) {$X_8$};
    \node[obs] (X9) at (4.8, 1) {$X_9$};
    \node[obs] (X10) at (7.2, 3) {$X_{10}$};

    \node[intv] (S1) at (0.0, 3) {$\sigma_1$};
    \node[intv] (S2) at (2.4, 3) {$\sigma_2$};
    \node[intv] (S3) at (4.8, 3) {$\sigma_3$};
    \node[intv] (S4) at (1.2, 1) {$\sigma_4$};
    \node[intv] (S5) at (8.4, 0) {$\sigma_5$};
    \node[intv] (S6) at (1.2, 0) {$\sigma_6$};
    \node[intv] (S7) at (8.4, 2) {$\sigma_7$};
    \node[intv] (S8) at (8.4, 1) {$\sigma_8$};
    \node[intv] (S9) at (4.8, 0) {$\sigma_9$};
    \node[intv] (S10) at (8.4, 3) {$\sigma_{10}$};

    \edge[black]{X2}{X1};
    \edge[black]{X2}{X3};
    \edge[black]{X3}{X4};
    \edge[black]{X3}{X5};
    \edge[black]{X3}{X6};
    \edge[black]{X3}{X7};
    \edge[black]{X8}{X5};
    \edge[black]{X8}{X7};
    \edge[black]{X9}{X4};
    \edge[black]{X9}{X5};
    \edge[black]{X10}{X7};

    \edge[black]{S1}{X1};
    \edge[black]{S2}{X2};
    \edge[black]{S3}{X3};
    \edge[black]{S4}{X4};
    \edge[black]{S5}{X5};
    \edge[black]{S6}{X6};
    \edge[black]{S7}{X7};
    \edge[black]{S8}{X8};
    \edge[black]{S9}{X9};
    \edge[black]{S10}{X10};
    
  \end{tikzpicture}%
  \\
  \vspace{0.5cm}
  \\
  \begin{tikzpicture}

    \hspace{0.15in}
    \node[right] (L) at (-1, 2) {\textbf{b.}};

    \node[obs] (X1) at (0.0, 0) {$X_1$};
    \node[obs] (X2) at (1.2, 0) {$X_2$};
    \node[obs] (X3) at (2.4, 0) {$X_3$};
    \node[obs] (X4) at (3.6, 0) {$X_4$};
    \node[obs] (X5) at (4.8, 0) {$X_5$};
    \node[obs] (X6) at (6.0, 0) {$X_6$};
    \node[obs] (X7) at (7.2, 0) {$X_7$};
    \node[obs] (X8) at (8.4, 0) {$X_8$};
    \node[obs] (X9) at (9.6, 0) {$X_9$};
    \node[obs] (X10) at (10.8, 0) {$X_{10}$};

    \node[fact] (F1) at (0.0, 3) {\textcolor{white}{$f_1$}};
    \node[fact] (F2) at (1.2, 3) {\textcolor{white}{$f_2$}};
    \node[fact] (F3) at (2.4, 3) {\textcolor{white}{$f_3$}};
    \node[fact] (F4) at (3.6, 3) {\textcolor{white}{$f_4$}};
    \node[fact] (F5) at (4.8, 3) {\textcolor{white}{$f_5$}};
    \node[fact] (F6) at (6.0, 3) {\textcolor{white}{$f_6$}};
    \node[fact] (F7) at (7.2, 3) {\textcolor{white}{$f_7$}};
    \node[fact] (F8) at (8.4, 3) {\textcolor{white}{$f_8$}};
    \node[fact] (F9) at (9.6, 3) {\textcolor{white}{$f_9$}};
    \node[fact] (F10) at (10.8, 3) {\textcolor{white}{$f_{10}$}};

    \node[intv] (S1) at (0.0, 4) {$\sigma_1$};
    \node[intv] (S2) at (1.2, 4) {$\sigma_2$};
    \node[intv] (S3) at (2.4, 4) {$\sigma_3$};
    \node[intv] (S4) at (3.6, 4) {$\sigma_4$};
    \node[intv] (S5) at (4.8, 4) {$\sigma_5$};
    \node[intv] (S6) at (6.0, 4) {$\sigma_6$};
    \node[intv] (S7) at (7.2, 4) {$\sigma_7$};
    \node[intv] (S8) at (8.4, 4) {$\sigma_8$};
    \node[intv] (S9) at (9.6, 4) {$\sigma_9$};
    \node[intv] (S10) at (10.8, 4) {$\sigma_{10}$};

    \edge[black]{S1}{F1};
    \edge[black]{S2}{F2};
    \edge[black]{S3}{F3};
    \edge[black]{S4}{F4};
    \edge[black]{S5}{F5};
    \edge[black]{S6}{F6};
    \edge[black]{S7}{F7};
    \edge[black]{S8}{F8};
    \edge[black]{S9}{F9};
    \edge[black]{S10}{F10};

    \uedge[black]{X1}{F1};
    \uedge[black]{X2}{F1};
    \uedge[black]{X2}{F2};
    \uedge[black]{X3}{F3};
    \uedge[black]{X2}{F3};
    \uedge[black]{X4}{F4};
    \uedge[black]{X3}{F4};
    \uedge[black]{X9}{F4};
    \uedge[black]{X5}{F5};
    \uedge[black]{X3}{F5};
    \uedge[black]{X8}{F5};
    \uedge[black]{X9}{F5};
    \uedge[black]{X6}{F6};
    \uedge[black]{X3}{F6};
    \uedge[black]{X7}{F7};
    \uedge[black]{X3}{F7};
    \uedge[black]{X8}{F7};
    \uedge[black]{X10}{F7};
    \uedge[black]{X8}{F8};
    \uedge[black]{X9}{F9};
    \uedge[black]{X10}{F10};
    
  \end{tikzpicture}
  
  \end{tabular}
 }
 \caption{The DREAM structural assumptions, following a process analogous to the Sachs et al. case described in Figure~\ref{fig:graph_sachs}.}
 \label{fig:graph_dream}
\end{figure}

The second simulator is the causal IFM. 
The factorization comes from the structure of the DAG, using the moralization criterion described in the previous section. 
Figures \ref{fig:graph_sachs}\textbf{(b)} and \ref{fig:graph_dream}\textbf{(b)} show the respective IFM graph structure. 
This results in the following factorization form:
\begin{equation}
p(x; \sigma) \propto \prod_{k = 1}^l f_k(x_{\{k\}} \cup \textbf{pa}(x_k); \sigma_{F_k}),
\label{eq:causal_chain_graph}
\end{equation}
\noindent where $\textbf{pa}(x_k)$ comes from the respective theoretical causal DAG case, with $F_k$ given accordingly by $k$. 

To learn the causal IFM simulator, we use pseudo-likelihood and assign each factor again to a black-box MLP of 15 hidden units where the corresponding $\sigma_{F_k}$ is a switch between independent sets of parameters within each factor. 
We additionally perform a discretization step for variable $X_i$ by collecting all data and doing uniformly binning it in 20 bins, so that it is faster to compute the conditional normalizing constants\footnote{While it is theoretically possible to use continuous variables and automatic differentiation through a quadrature method that computes each univariate integral for each term in the pseudo-likelihood, this is still far too slow in practice. The discretization level chosen for these examples are fine enough so that it does not appear to affect the predictive performance of the $p(y~|~x)$.} for each term in the pseudo-likelihood objective function.

To sample from the learned IFM, so that we can numerically compute quantities such as $\mu_{\sigma}$,  we use Gibbs sampling.

\vspace{10pt}

\subsection{Generating Ground-Truth Population Models and Data}
\label{appendix:generate_x_and_y}

Generating ground truth data, either for numerically computing population quantities by Monte Carlo or as a generator of training data, includes the following steps: (1) learning simulators, (2) generating ground truth $X$ and (3) generating ground truth $Y$.

\paragraph{Learning simulators.}

The first step involves learning the simulators: with the Sachs et al.~data, we use 5 data regimes as the training data for the simulator (1 baseline regime and 4 interventional); and with DREAM data, we use 11 data regimes as the training data for the simulator (1 baseline regime and 10 interventional). As described above, two simulators are built for each of the two studies.


\paragraph{Generating ground-truth system $X$.}

Since each intervention is considered as a binary value (0 for no intervention and 1 for with intervention), with the training dataset of 5 data regimes in Sachs, this gives us a total of $2^4=16$ combinatorial possibility regimes; as for the DREAM case, we have in total of 11 regimes, which means that the complete space $\Sigma$ has $2^{10} = 1024$ combinations. To simplify the computation of the benchmark, we are interested in the "one-to-double knockdown" scenario and hence generate a total of 56 regimes ($= \frac{10 \times 9}{2}+11$).

The original training datasets for both simulators are discarded and we now consider the simulator as the oracle for any required training set and population functionals. In particular, for each regime, we generate 25,000 samples to obtain a Monte Carlo representation of the ground-truth respective population function $p(x; \sigma)$. 

\paragraph{Generating ground-truth outcome processes $Y$.}

For outcome variables $Y$ from which we want to obtain $\mu_{\sigma^\star} := \mathbb E[Y; \sigma^\star]$ for given test regimes $\sigma^\star$, we consider models of the type $Y = \tanh(\lambda^\top X) + \epsilon_y$, with random independent normal weights $\lambda$ and $\epsilon_y \sim \mathcal{N}(0, v_y)$. $\lambda$ and $v_y$ are scaled such that the ground truth variance of $\lambda^{\mathsf T}X$ is a number $v_x$ sampled uniformly at random from the interval $[0.6, 0.8]$, and set $v_y := 1 - v_x$. 

For each of the four benchmarks (i.e., based on either the Sachs et al. data or DREAM data, with either a DAG model-based ground-truth or an IFM-based ground truth), we generate 100 random vectors $\lambda$. The point of these 100 problems is just to illustrate the ability to learn (noisy) summaries of $X$, or general downward triggers or markers predictable from $X$ under different conditions. When generating $Y$ from $X$, \emph{we keep a single sample for $X$}. We then generate an unique sample for each of the 100 $Y$ variations given the same $X$ data.

\paragraph{Generating training data.}
For training our models, we additionally generate 5000 samples for the observational regime (baseline) and 500 samples for each of the remaining 4 experimental conditions (Sachs) and 10 experimental conditions (DREAM). To map from $X$ to $Y$, we use the model described above.

\section{Further Experimental Results}
\label{appendix:further_results}



\begin{figure}[h]
  \centering
  \includegraphics[width=1\textwidth]{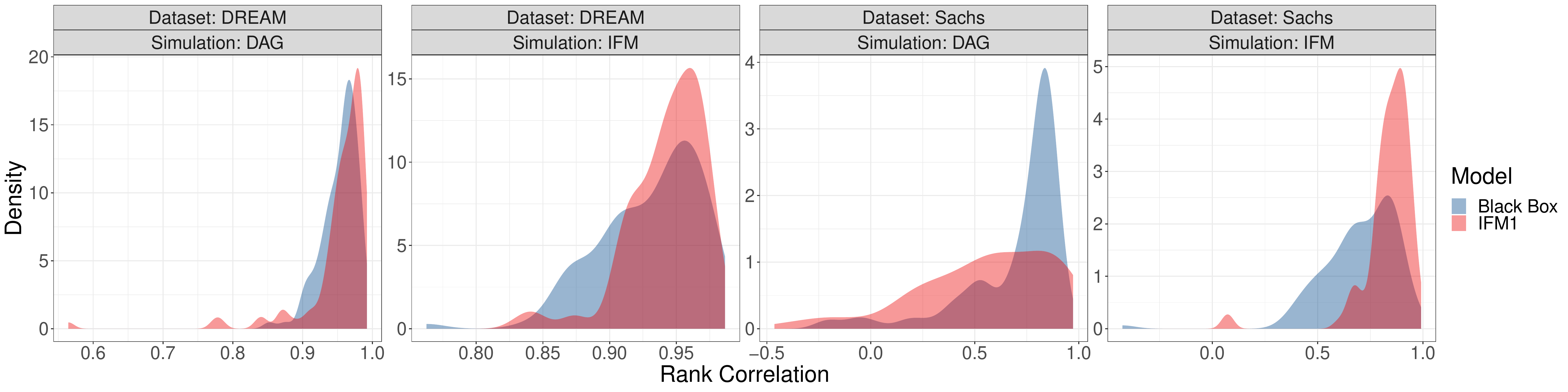}
  \caption{Overlapping density plots showing average rank correlation between true treatment effects and those predicted by the black box model and IFM1, respectively. Ideal performance is a point mass on $1$.}
  \label{fig:hist}
\end{figure}

In Table~\ref{tab:main_results} (see also Fig.~\ref{fig:hist})
we present a series of further experimental results in numerical form based on the following metrics: \textit{i)} \textit{proportional root mean squared error} (pRMSE): the average of the squared difference between the ground truth $Y$ and estimated $\widehat{Y}$, where each entry is further divided by the ground truth variance of the corresponding $Y$; 
and \textit{ii)} \textit{rank correlation} (rCOR): the Spearman's $\rho$ between the ground truth vector $\mu_{\sigma^\star}$ for all entries in $\Sigma_{\test}$, and the corresponding estimated vector.\footnote{The IFM-3 results are omitted from Tables~\ref{tab:main_results} and \ref{tab:pvalues}, as the method does not convergence in a reasonable time.}

In Table~\ref{tab:pvalues}, 
we present results from a series of one-sided binomial tests to determine whether models significantly outperform the black box baseline. 


\begin{table}[h]
\scriptsize
\centering
\caption{Results of our interventional generalization experiments for the Sachs and DREAM datasets. 
The values correspond to the average of $100$ $Y$ problems.
}
\resizebox{.79\columnwidth}{!}{
\begin{tabular}{c|c|c|c|c}
\hline
 & \multicolumn{2}{c}{\textbf{Sachs}} & \multicolumn{2}{|c}{\textbf{DREAM}} \\
\hline
& \multicolumn{1}{c|}{\textit{Causal-DAG}} & \multicolumn{1}{c|}{\textit{Causal-IFM}} & \multicolumn{1}{c|}{\textit{Causal-DAG}} & \multicolumn{1}{c}{\textit{Causal-IFM}} \\
\hline
& \multicolumn{1}{c|}{\textbf{pRMSE}} & \multicolumn{1}{c|}{\textbf{pRMSE}} & \multicolumn{1}{c|}{\textbf{pRMSE}} & \multicolumn{1}{c}{\textbf{pRMSE}} \\
\hline
Blackbox & $0.043$  & $0.414$  & $0.025$  & $0.174$  \\
Causal-DAG & $0.014$ & $0.408$  & $0.017$  & $1.337$  \\
IFM-1 & $0.105$  & $0.168$  & $0.022$  & $0.185$  \\
IFM-2 & $0.051$  & $0.111$  & $0.107$  & $0.769$ \\
IFM-3 & $0.123$ & $0.175$  & -- & --  \\
\hline
& \multicolumn{1}{c|}{\textbf{rCOR}} & \multicolumn{1}{c|}{\textbf{rCOR}} & \multicolumn{1}{c|}{\textbf{rCOR}} & \multicolumn{1}{c}{\textbf{rCOR}} \\
\hline
Blackbox & $0.696$  & $0.701$ & $0.953$  & $0.930$  \\
Causal-DAG & $0.873$  & $0.405$  & $0.972$  &  $0.502$  \\
IFM-1 & $0.546$  & $0.835$  & $0.952$  & $0.942$ \\
IFM-2 & $0.673$ & $0.821$  & $0.865$  & $0.737$  \\
IFM-3 & $0.503$  & $0.811$  & -- & -- \\
\hline
\end{tabular}}
\label{tab:main_results}
\end{table}


\newpage
\begin{table}[h]
\small
\centering
\caption{\mbox{P-values} from a series of one-sided binomial tests against the null hypothesis that models perform no better on average than the black box model. Significance at \mbox{$\alpha = 0.05$} is indicated with one asterisk, and \mbox{$\alpha = 0.001$} with two.}
\label{tab:pvalues}
\begin{tabular}{llrrrr}
  \toprule
Data  & Simulation & DAG   & IFM1  & IFM2  & IFM3 \\
  \midrule
DREAM & Causal-DAG & $< 0.001^{**}$     & $0.044^*$ & $1$     & NA   \\
DREAM & Causal-IFM & $1$     & $0.972$ & $1$     & NA   \\
Sachs & Causal-DAG & $< 0.001^{**}$     & $1$     & $0.998$ & $1$    \\
Sachs & Causal-IFM & $0.956$ & $< 0.001^{**}$     & $< 0.001^{**}$     & $< 0.001^{**}$   \\
  \bottomrule
\end{tabular}
\end{table}
\vspace{20pt}

\end{document}